\documentclass{article}

\usepackage{PRIMEarxiv}
\usepackage{amsmath}
\usepackage[utf8]{inputenc} 
\usepackage[T1]{fontenc}    
\usepackage{hyperref}       
\usepackage{url}            
\usepackage{booktabs}       
\usepackage{amsfonts}       
\usepackage{nicefrac}       
\usepackage{microtype}      
\usepackage{lipsum}
\usepackage{fancyhdr}       
\usepackage{graphicx}       
\usepackage{xspace}
\newcommand{\algo}{FETILDA\xspace}
\graphicspath{{media/}}     

\title{\algo: An Effective Framework For Fin-tuned Embeddings For Long Financial Text Documents
}

\author{
  Bolun ``Namir'' Xia \thanks{These authors contributed equally to this work.}\\
  Department of Computer Science \\
  School of Science \\
  Rensselaer Polytechnic Institute \\
  Troy, NY, USA\\
  \texttt{xiabolun@gmail.com} \\
   \And
  Vipula D. Rawte \footnotemark[1]\\
  Department of Computer Science \\
  School of Science \\
  Rensselaer Polytechnic Institute \\
  Troy, NY, USA\\
  \texttt{rawtevipula25@gmail.com} \\
   \And
  Mohammed J. Zaki \\
  Department of Computer Science \\
  School of Science \\
  Rensselaer Polytechnic Institute \\
  Troy, NY, USA\\
  \texttt{zaki@cs.rpi.edu} \\
   \And
  Aparna Gupta \\
  Lally School of Management \\
  Rensselaer Polytechnic Institute \\
  Troy, NY, USA\\
  \texttt{guptaa@rpi.edu} \\
}

\begin{document}
\maketitle

\begin{abstract}
Unstructured data, especially text, continues to grow rapidly in
various domains. In particular, in the financial sphere, there is a wealth
of accumulated unstructured financial data, such as the textual
disclosure documents that companies submit on a regular basis to regulatory
agencies, such as the Securities and Exchange Commission (SEC). These
documents are typically very long and tend to contain valuable soft
information about a company's performance, which is usually not captured in
predictive analyses that only utilize quantitative data. It is therefore of
great interest to learn predictive models from these long textual documents,
especially for forecasting numerical key performance indicators (KPIs).
Whereas there has been a great progress in natural language processing via
pre-trained language models (LMs) learned from tremendously large corpora of
textual data, they still struggle in terms of effective representations for
long documents. Our work fills this critical need, namely how to develop
better models to extract useful information from long textual documents and
learn effective features that can leverage the soft financial and risk
information for text regression (prediction) tasks. In this paper, we
propose and implement a deep learning framework that splits long documents
into chunks and utilizes pre-trained LMs to process and aggregate the chunks
into vector representations, followed by attention-based sequence modeling
to extract valuable document-level features. We evaluate our model on a
collection of 10-K public disclosure reports submitted annually by US banks,
and another dataset of reports submitted by US companies. Overall, our framework outperforms
strong baseline methods for textual modeling as well as a baseline
regression model using only numerical data.
Our work provides better insights into how utilizing
pre-trained domain-specific and fine-tuned long-input LMs in representing
long documents can improve the quality of representation of textual data,
and therefore, help in improving predictive analyses.
\end{abstract}


\section{Introduction}\label{sec1}
Unstructured data such as text is growing very fast in different domains.
Especially, textual data from financial documents has been found to be
beneficial in making predictions \cite{das2014text}. Utilizing such large
volumes of textual data requires natural language processing (NLP) and
machine learning (ML) techniques. These techniques summarize the text as (a
set of) numeric feature vectors, which are called representations or
embeddings, and which can in turn serve as inputs to machine learning models
to predict some target variables.

The traditional approach for text-based learning is via the Term Frequency -
Inverse Document Frequency (TF-IDF) method~\cite{Jurafsky21}, which can
represent the document as a long numeric vector of TF-IDF scores for each
word. However, TF-IDF does not attempt to directly extract the latent
semantic information within the text. The current progress in text
representations was initiated by word embedding methods, such as word2vec
\cite{mikolov2013efficient} and GloVe \cite{pennington2014glove}, which
capture both the lexical and semantic information of a document to some
extent. The main idea is to learn word representations based on the context
of each word. However, these methods learn only a single, \textit{static}
representation for each word, and do not take into consideration the
phenomenon of polysemy, where a word can change its meaning depending on the
context (for example, the the word `bank' in the financial context has a
very different meaning compared to the `bank' of a river). The
state-of-the-art (SOTA) pre-trained language models, such as GPT
\cite{GPT-1} and BERT \cite{BERT} are built on top of the very effective
Transformer-based attention model~\cite{vaswani2017attention}, which can
learn \textit{contextual} word embeddings. These embeddings are \textit{dynamic} 
in terms of the surrounding block or context of the word, so that
the same word can get different representations that are most effective in
capturing the lexical and semantic information. These models have shown SOTA
performance on a variety of downstream tasks such as question answering,
text classification, and regression.

While much progress has been made in NLP, our focus in this paper is on the
relatively under-examined area of financial text documents. Specifically, we
are focusing on the 10-K financial reports that companies in the US submit
annually to the Securities and Exchange Commission (SEC). These reports
describe a company’s activities, progress, and risks in the interest of the
company's stakeholders. The detailed content of these reports is helpful in
evaluating the status of the company as well as predicting future metrics
from forward-looking statements. One of the key parts of 10-K reports is
Item 7, namely, the Management's Discussion and Analysis (MDA) section.
Another key part is Iten 1A, namely, the Risk Factors section. Textual data
contained in these 10-K reports are predictive of the volatilities that
companies have in the stock market, and can be helpful in predicting
failures of financial institutions, as well
\cite{das2014text,loughran2016textual}. But, in these approaches, they use
an expert-generated sentiment dictionary by Loughran and McDonald (L\&M)
\cite{loughran2011liability} to generate the vector of features. For
example, Kogan et al. \cite{kogan2009predicting} utilized the MDA section to
predict the return volatility of stocks using the L\&M word list. In
addition, the L\&M word list was expanded using word2vec
\cite{mikolov2013efficient}. Then, it was used in Tsai and Wang
\cite{tsai2017risk}.

As mentioned above, moving beyond the static embedding methods, it is
important that we take into consideration the contextual information of
words, since specific words can have different meanings in the financial
context than those in the general context, the nuances of which may not be
apparent to laypersons. For this, we can utilize the recent contextual
language models to represent long documents, such as the 10-K reports, in
order to construct effective models for conducting predictive analyses.
However, extracting ``good'' representations for such long documents remains
a challenging task: the length of the 10-K documents poses both a
methodological and ontological burden. Methodologically, financial reports
are significantly longer, compared to the maximum length of a textual
sequence that BERT \cite{BERT}-based models can handle. For instance, the
Management Discussion and Analysis (MD\&A) section of the 10-K reports that
companies publish annually is usually around 12,000 word-tokens. BERT-based
models have a restriction on the maximum number of tokens, around 512, with
some newer models, such as Longformer \cite{Longformer} and BigBird
\cite{BigBird}, reaching up to 4,096 tokens. Ontologically, the challenge is
the classic machine learning task of extracting or learning informative
features that can represent the input well. This question becomes quite
complex in the context of representing a long document. Contextual word
embeddings are well suited for this given their ability to ``understand''
different meanings for a word in different contexts. However, it remains an
open question as to how to combine the various contextual word embeddings
into an effective document level embedding.

An additional challenge is that the SOTA language models are pre-trained on
massive and generic corpuses, e.g., from web crawls, wiki media, and so on.
However, to be effective for the financial context, it is important for LMs
to learn domain-adapted and task-specific representations of long documents
in order to meaningfully support predictive analyses. This can usually be
achieved either by pre-training (from scratch) a domain-specific language
model on a huge financial corpus to adapt to its particular domain, or by
fine-tuning a pre-trained LM on a specific financial dataset for the
downstream tasks, or by combining the two approaches of adapting the LM to a
particular financial domain followed by fine-tuning on downstream tasks.
Recently there have been several attempts at pre-training BERT on large
financial corpuses to adapt it for tasks in the financial domain. Liu et
al.\cite{Liu}, Yang et al. \cite{Yang}, Araci \cite{Araci}, and DeSola et
al.\cite{desolafinbert}, each pre-trained the BERT model from scratch on
financial corpora, such as financial news, corporate reports, financial
websites, and so on. Incidentally, all four approaches are called FinBERT!


To address the long document representation challenge within the context of
financial disclosure documents, we propose a novel framework called  
{\bf F}in-tuned {\bf E}mbeddings of {\bf T}exts {\bf I}n {\bf L}ong {\bf
D}ocument {\bf A}nalysis (FETILDA).
Our approach is particularly designed for downstream predictive or
regression tasks, where the input document representations are combined with
other numeric attributes (if available) to predict a target response
variable of interest, such as key performance indicators (e.g., return on
assets, earnings per share), stock volatility, and so on. FETILDA comprises
a novel chunk-based deep learning framework, where a long document is split
into several smaller chunks and then each chunk is processed using an
appropriate language model (e.g., BERT~\cite{BERT}, FinBERT \cite{Yang} or
Longformer \cite{Longformer}). The layers of the LM can remain frozen, or
they may be unfrozen for fine-tuning, or only the last layer can be frozen.
The chunk level representations are then pooled together using a Bi-LSTM
model equipped with  self-attention mechanism. The pooled chunks are then
aggregated into a document level representation, which serves as input for a
fully connected linear neural network for target variable prediction. We can
also use the document representations as inputs to  Support Vector and
Kernel regression models. We evaluate our framework using two different
corpuses: i) 10-K reports submitted annually to the SEC by US banks for the
period from 2006 to 2016, ii) 10-K reports for all US companies from 1996 to
2013 \cite{FIN10K}. We have conducted extensive experiments using these
datasets and applied our framework to different predictive analysis
regression tasks: i) predictive metrics of a bank's financial performance,
such as Return on Assets (ROA), Earnings Per Share (EPS), Return on Equity
(ROE), Tobin's Q Ratio (TQR), Leverage Ratio (LR), Tier 1 Capital Ratio
(T1CR), Z-Score (Z) and Market to Book Ratio (MBR); ii) analysis of a
company's stock market volatility. Our results compared against the
different baseline methods show that FETILDA performs significantly better
and yields SOTA results for long financial text regression tasks. In
summary, our main contributions are:

\begin{itemize}

\item We propose a SOTA deep learning framework for long document regression
    tasks in the financial domain. Our FETILDA approach is designed to learn
    effective document level representations via a sequential chunking
    approach combined with an attention mechanism. As such our approach
    combines the best of both the attention-based Transformer model and
    Bi-LSTM recurrent networks.

\item We conduct an extensive set of experiments to quantitatively showcase
    the effectiveness of our FETILDA approach. We applied the model on two
    different 10-K datasets, and on 9 different regression tasks. We show
    that FETILDA outperforms several different baseline methods, and
    achieves state-of-the-art results on long financial documents.

\end{itemize}



\section{Related Work}\label{sec2}
Machine learning plays an important role in financial analytics. One of the
important areas of finance is investment stock return forecasting, as well
as fundamentals forecasting and risk modeling, that mainly employ
quantitative or numeric data \cite{emerson2019trends}. Different ML models
such as Support Vector Machines (SVMs), single hidden layer Feed-forward
Neural Networks and Multi-layer Perceptrons (MLPs) were used for the
prediction of future price movements in \cite{nousi2019machine}. They mainly
used two sets of features for their ML classifiers: (1) handcrafted features
formed on the raw order book data and (2) features extracted using ML
algorithms. Some other models such as Random Forest
\cite{khaidem2016predicting}, XGBoost \cite{wang2019xgboost}, Bidirectional
Long Short-Term Memory (Bi-LSTM) and stacked LSTMs
\cite{sardelich2018multimodal} were also implemented to predict business
risk and stock volatility. The main limitation of these works is that they
ignore valuable textual data that can provide more insight into the
intangible features such as sentiment, knowledge capital, risk culture, and
so on. 

\subsection{Textual Data: Sentiment Analysis}
An approach to predict financial quantitative variables is using financial
textual sources such as news reports, analyst assessments, earnings call
transcripts, and company filing reports. 
In \cite{tetlock2008more}, textual features were created by using the
negative words in the Harvard-IV-4 TagNeg dictionary and constructing a
document-term matrix from the news stories. These features were used to
predict firms’ earnings and stock returns. A novel tree-structured LSTM was
proposed to automatically measure the usefulness of financial news using
both text and cumulative abnormal returns \cite{chang2016measuring}. A
dual-layer attention-based neural network model was developed to predict
stock price movement using the text in financial news
\cite{yang2018explainable}. Estimating the value of text in financial news
is important because it drives the investment decision making process. 

Financial sentiment analysis is challenging because of lack of labeled data
specific to this domain. Moreover, the general-purpose pre-trained language
models fail to capture the financial context. \cite{Araci} proposed the
FinBERT model, which can be fine-tuned on the financial sentiment analysis
dataset (FiQA) to outperform the general BERT model. Besides financial news,
in \cite{kogan2009predicting}, the authors constructed textual features from
10-K reports. They used these features to predict the future stock
volatility indicating the effectiveness of text. A deep learning model
trained on the SEC filings was used to improve the prediction of company’s
stock price over the traditional ML models \cite{sakarwala2019use}. 

The authors in \cite{tsai2016discovering} and \cite{tsai2017risk} extracted
additional textual features by expanding the L\&M sentiment word
list~\cite{loughran2011liability} semantically and syntactically, using
word2vec \cite{mikolov2013efficient}. Similarly, the uncertainty word list
in L\&M dictionary was expanded using word2vec to predict stock volatility
\cite{theil2018word}. The authors in \cite{theil2020explaining} expanded the
L\&M dictionary by training industry-specific word embedding models using
word2vec to predict volatility, analyst forecast error and analyst
dispersion. \cite{sedinkina2019automatic} showed how automatic domain
adaption of the L\&M sentiment list using word2vec
\cite{mikolov2013efficient} improved the prediction of excess return and
volatility. The aforementioned dictionary expansion approaches used word2vec
model to select the top $k$ closest words to the words existing in the L\&M
dictionary. Since word2vec is a model based on static word embeddings, it
fails to capture the dynamic context of the words. 

\subsection{Language Models in Finance}
In terms of domain-adapted pre-trained LMs, in the English-speaking Finance
sphere, four models have been proposed and implemented, all named FinBERT:
Liu et al. \cite{Liu}, Yang et al. \cite{Yang}, Araci \cite{Araci}, and
DeSola et al. \cite{desolafinbert}, all of which are pre-trained to adapt to
different financial domains. Originally, in the general domain, BERT
\cite{BERT} was pre-trained on two corpora: BooksCorpus (0.8 billion words),
and English Wikipedia (2.5 billion words), forming a total of 3.3 billion
words, so the idea of these financial language models is to take the
original model, and pre-train it on their respective financial corpora.

Araci \cite{Araci} was the first to propose FinBERT as a pre-trained
domain-adapted BERT \cite{BERT} on a corpus called TRC2-financial, which
includes 46,143 documents with more than 29M words and nearly 400K
sentences, from a set of Reuters news stories. 
In experimentation, they saw a 15\% increase in accuracy for classification
tasks, a significant margin. Liu et al. \cite{Liu} focused on financial news
and dialogues present on websites, and collected three financial corpora: 13
million financial news (15GB) and financial articles (9GB) from Financial
Web, totaling 24GB and 6.38 billion words; financial articles from Yahoo!
Finance, totaling 19GB and 4.71 billion words; and question-answer pairs
about financial issues from Reddit, totaling 5GB and 1.62 billion words.
They pre-trained their model on these corpora to adapt it to the financial
news and dialogues domain. In experimentation, they saw their model
outperform BERT \cite{BERT} on all the financial tasks in their experiments,
in terms of accuracy, precision, and recall \cite{Liu}. Yang et al.
\cite{Yang} focused on financial and business communications that companies
produce, and collected a corpus of three types of data: 10-K and 10-Q
reports, totaling 2.5 billion word tokens; earnings call transcripts,
totaling 1.3 billion word tokens; and analyst reports, totaling 1.1 billion
word tokens \cite{Yang}. 
They report that their model outperforms BERT \cite{BERT} in three sentiment
analysis tasks, all by significant margins \cite{Yang}. DeSola et
al.~\cite{desolafinbert} introduced another domain-specific pre-trained
language model, also named FinBERT, for financial NLP applications. This
model was trained on the 10-K filings from 1998 to 1999 and from 2017 to
2019, totaling 497 million words, and it showed better performance than BERT
on the masked LM and next sequence prediction tasks.

\subsection{Long Document Language Models}
Apart from LMs adapted to specialized domains, there has been a slew of
papers on state-of-the-art pre-trained LMs in the general domain, such as
GPT-1 \cite{GPT-1}, GPT-2 \cite{GPT-2}, GPT-3 \cite{GPT-3}, T-5 \cite{T-5},
ELECTRA \cite{ELECTRA}, and so on. These are massive models trained on
enormous corpora, but the challenge of representing long documents persists,
in that these models still cannot handle long textual sequences, due to the
quadratic computational complexity that they usually entail.

To tackle this challenge head-on, several recent works, such as Longformer
\cite{Longformer}, ETC \cite{ETC}, and BigBird \cite{BigBird}, have been
proposed, all of which innovate on the self-attention mechanism in order to
reduce the computational complexity from quadratic to linear, which then
enables it to process longer sequences of text. In addition, more recent
works on transformer models with linear attention, such as Reformer
\cite{kitaev2020reformer} and Nystromformer \cite{xiong2021nystromformer},
innovate on how to mathematically approximate the self-attention matrix
calculations with less time and space complexity, instead of changing the
self-attention mechanism.

Longformer \cite{Longformer} replaces the full self-attention matrix, which
scales quadratically with the length of the input sequence, with three types
of sparse attention schemes: sliding window attention, which selects only
the entries on the descending diagonal line of the self-attention matrix,
with the `thickness' of the line being a certain size; dilated window
attention, which adds gaps of a certain size in between the sliding window,
making the descending diagonal line dilated; global attention, which has
certain specific tokens attend to all the tokens across the sequence, both
horizontally and vertically, thereby enabling global contextual
representation of the sequence. Longformer was shown to outperform baseline
methods consistently, and particularly, its results were more apparent where
the experiment required long contextual information.

Extended Transformer Construction (ETC) \cite{ETC} is very similar to
Longformer, with nuanced variations. ETC replaces the full self-attention
matrix with global-local attention, which splits the self-attention matrix
into four parts: global-to-global, which is a small square on the top left
of the matrix, where certain special global tokens attend to each other;
global-to-long, which is a horizontal rectangle on the top right of the
matrix, where global tokens attend to regular tokens; long-to-global, which
is a vertical rectangle on the bottom left of the matrix, where regular
tokens attend to global tokens; long-to-long, which is a compressed version
of the descending diagonal line in the large square on the bottom right of
the matrix, essentially a sliding window attention compressed into a
rectangular matrix, where regular tokens attend to other regular tokens in
its window. In experimentation, ETC yielded state-of-the-art results,
especially in question answering scenarios.

BigBird \cite{BigBird} extends further on ETC \cite{ETC}, adding random
sparse attention into the mix, building on top of the global-local attention
mechanism of ETC. Random entries in the self-attention matrix are selected
to generalize over the full matrix. From a graph theory perspective, this
means a shorter average path between any two nodes, making it a better
approximation of the full graph. And from a NLP perspective, in most texts,
there tends to be locality of reference, where a word relates closely with
words around it, so BigBird tries to account for this with their particular
random sparse attention scheme.

\subsection{Chunking-based Representation Schemes}
Similar to the \algo approach we propose in this paper, other
approaches to constructing document-level representations have been proposed
in other works. Yang et al.\ \cite{YangLiu} proposed a method similar to ours in their
hierarchical approach of using sentence blocks to construct document
representations, for the purpose of document matching. In their approach,
the sentence block representations are passed through a
transformer model and the first token of the resulting output is used as the
document representation. They also have an option to concatenate
the sentence representations via attention based weights.
With this framework, they are able to handle maximum document
lengths ranging from 512 to 2,048 word tokens. However, in our use case,
\algo has to deal with with much longer documents than 2,048 word
tokens. Even just one section of the 10-K reports that we learn on, namely,
Item 7, averages around 12,000 words, and can be more than 24,000
tokens after tokenization. Therefore, instead of sentence-level blocks,
we use (paragraph-level) chunks, and we weigh these chunks using their
respective attention scores obtained by a self-attention mechanism. 
Finally, we pool the weighted chunks together into a document
representation, using a bi-LSTM network. In our experiments, we utilize
different models that can enable us to have the maximum chunk length range
from 512 up to 8,192 word tokens, which then enables us to generate
representations for documents that average 24,000 word tokens. Furthermore,
while both our approach and their approach focus on obtaining a good
document representation, the final goals are different: we aim to learn
financial text features predictive of future target metrics of banks and
companies, while their aim is document matching.

Gong et al.\ \cite{gong-etal-2020-recurrent} proposed a recurrent chunking
mechanism for the purpose of machine reading comprehension, where the
machine is given a long document and a question, and is required to extract
a piece of text from the document as the answer to the questions. Towards
that end, they needed the chunking mechanism to be such that the separation
point of various chunks would not cut the correct answer in half, nor
prevent surrounding contexts from being retained. Therefore, their main
innovation is in enabling a more flexible chunking policy, and in a
recurrent chunking mechanism that can provide context surrounding a chunk
segment. In experimentation, they use BERT, which enables maximum sequence
lengths ranging from 192 to 512 word tokens. Our framework 
takes a different approach to the chunking policy, 
since our goal is the extraction of 
predictive financial text features from the document. Moreover, we also try
to increase the maximum sequence length of a chunk up to 8,192 word tokens, 
which then allows us to take in more information in one chunk
in an organic way. 

In more recent work, Grail et al. \cite{grail-etal-2021-globalizing} 
use a bi-GRU network to
pool the chunks together, instead of a bi-LSTM network.
But the purpose
of their framework is long document summarization, instead of extracting
predictive text features. In their approach, they use BERT as the LM, and
therefore, can only process up to 512 word tokens for a chunk. Further, they
consider their approach to be an alternative to long sequence LMs such as
Longformer. Instead of considering long sequence LMs as
alternatives, we incorporated them into our FETILDA framework. Therefore, in
our approach, we experiment with various underlying LMs for FETILDA, be it
BERT-based models, or long sequence LMs, or linear attention transformer
LMs, enabling us to have different maximum sequence lengths ranging from 512
to 8,192 tokens.

\begin{figure}[!ht]
\centering
  \includegraphics[scale=0.41]{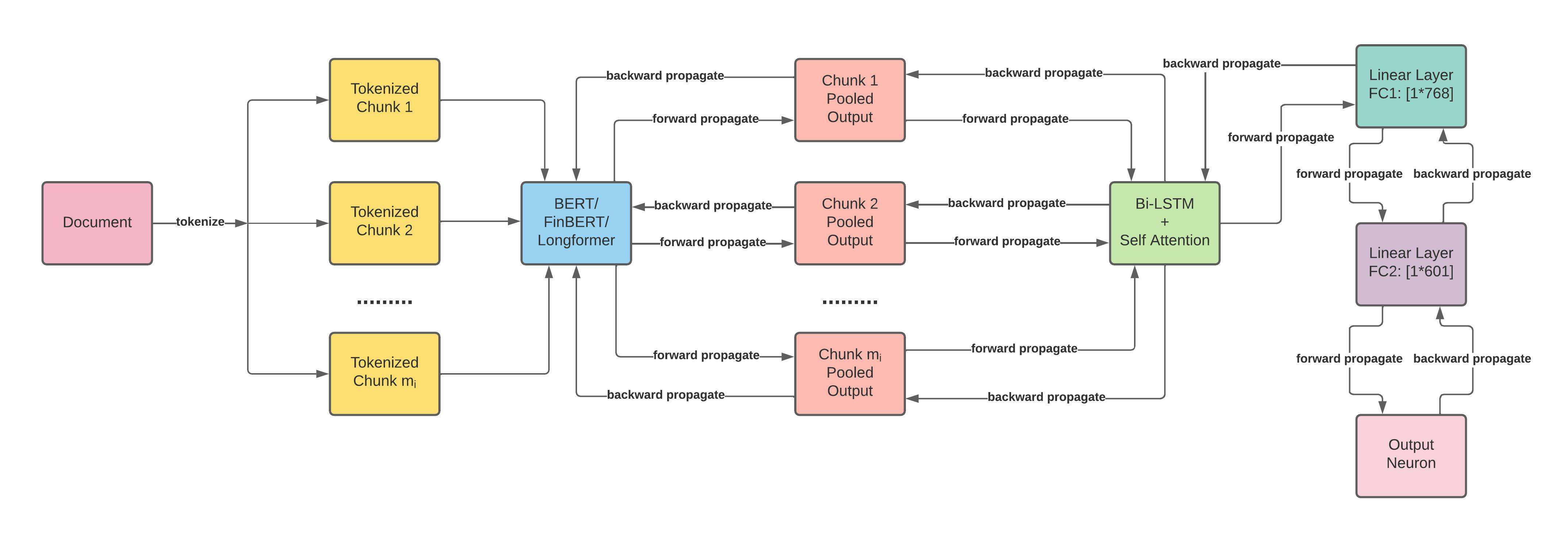}
  \caption{\algo: Overall Framework.}
  \label{fig:overall}
\end{figure}

\section{\algo: Long Document Representation}
\label{sec3}

Figure~\ref{fig:overall} shows the complete \algo framework.
\algo first splits a long document into smaller fragments or chunks, then processes
each chunk using a language model, all of whose layers are fully unfrozen
for fine-tuning, then pools the chunks together using a Bi-LSTM layer
endowed with a self-attention mechanism into an aggregate vector
representation of the entire document. 
The chunk representations are extracted from the
underlying language model (BERT~\cite{BERT}, FinBERT~\cite{Yang}, Longformer~\cite{Longformer}, or Nystromformer~\cite{xiong2021nystromformer})  using several
different pooling strategies including using the default pooler output and
combining the features from the last few layers. These chunk sequences are
passed onto a Bi-LSTM model whose hidden context states and outputs are used
to learn chunk-level attention scores to extract the final document
embedding. Finally, the document embedding is passed through the linear
layers to obtain the final target prediction. In addition, we perform
task-specific fine-tuning on our entire model, including BERT, FinBERT, or
Longformer, whose layers are fully unfrozen (or can be kept frozen if only
pre-trained inputs are to be used), using MSE as the loss function.
Overall, as shown in  
Figure~\ref{fig:overall},
our methodology consists of
four stages: (1) Chunk Generation, (2) Chunk-Level LM Pooling, (3)
Document-Level Attention Pooling, and (4) Model Training and Fine-Tuning. We
shall describe each of these next.

\subsection{Chunk Generation}

\begin{figure}[!ht]
\centering
  \includegraphics[scale=0.51]{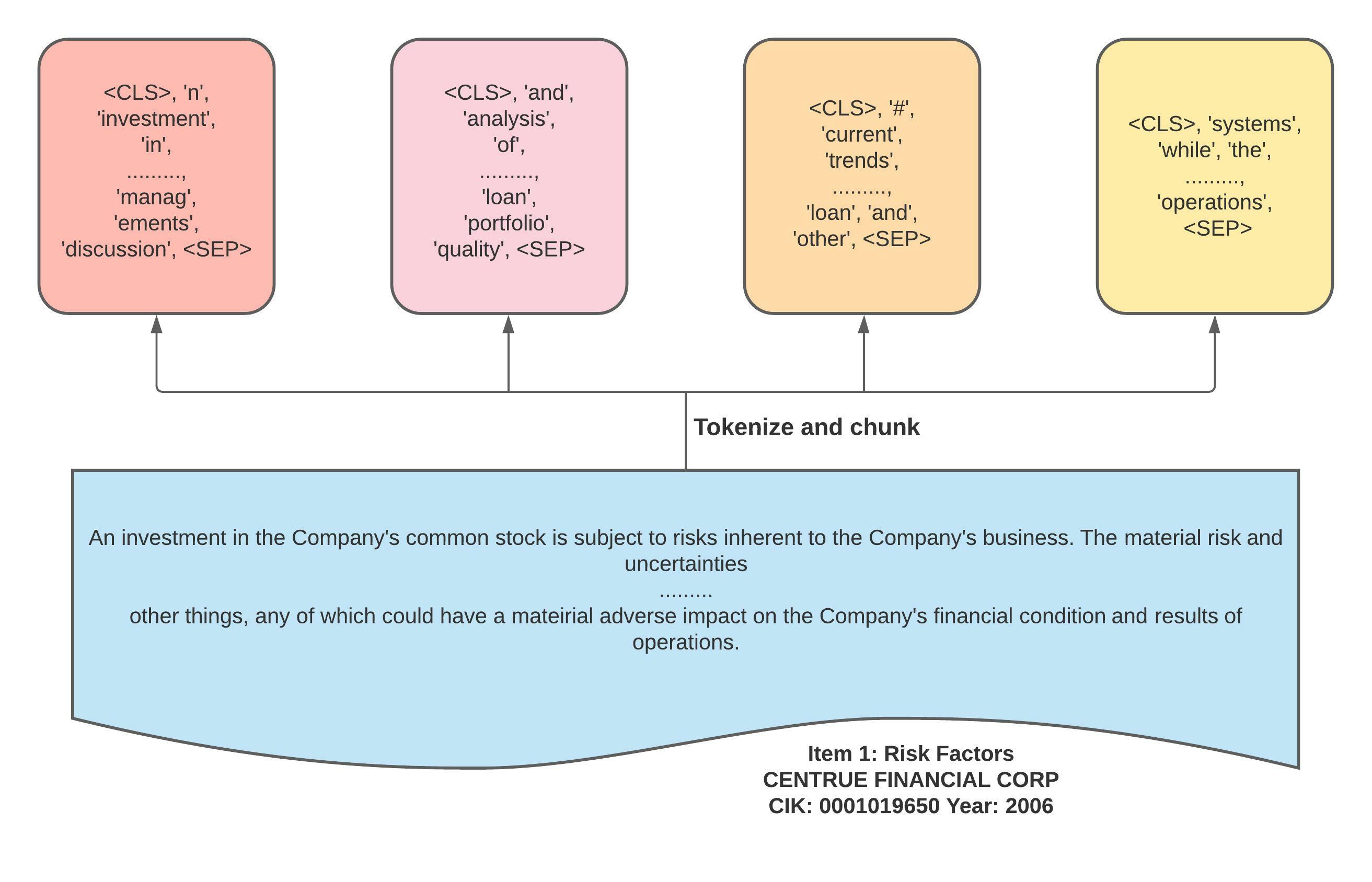}
  \caption{Chunk Generation}
  \label{fig:tokengen}
\end{figure}

Let $L = \{d_1, d_2, \cdots, d_N\}$ denote a text corpus containing $N$ long
documents, where $d_i$ denotes the $i$-th document in the corpus. We
tokenize each document $d_i$ into a sequence of tokens $\{t_1, t_2, \cdots,
t_{n_i}\}$, where $n_i$ is the number of tokens for document $d_i$. The
document token sequence is divided into chunks of length $b$, where $b$ is
the block or chunk size. Thus, each document $d_i$ can be represented as a
sequence of chunks $\{c_1, c_2, \cdots, c_{m_i}\}$, with $m_i$ chunks of length
$b$. We also prepend and append \texttt{<CLS>} and \texttt{<SEP>} tokens to
each chunk, respectively, resulting in chunks of size $b+2$. The chunk size
dictates a maximum of $b+2$ tokens for each chunk $c_i = \{t_0, t_1, \cdots,
t_b, t_{b+1}\}$, with $t_0 = \texttt{<CLS>}$ and $t_{b+1}= \texttt{<SEP>}$.
We experiment with $b+2=512$, $b+2=4096$ and $b+2=8192$, depending on the
underlying language model used.
For document where the last chunk has $k < b$ tokens, we pad the last chunk
by appending the padding token (\texttt{<PAD>}) $(b-k)$ times to keep the
chunk length intact. For each chunk, we also create an attention mask with
\texttt{[0]} for padding tokens and \texttt{[1]} for non-padding tokens,
which helps in attending only to the valid tokens and not the \texttt{<PAD>}
tokens. Figure~\ref{fig:tokengen} shows an excerpt from Item 1 from a
company's 10-K report, and the tokenization and chunking process with four
resulting chunks.

\subsection{Chunk-Level Language Model Pooling}\label{sec:3.2}
\begin{figure}[!htb]
\centering
  \includegraphics[scale=0.51]{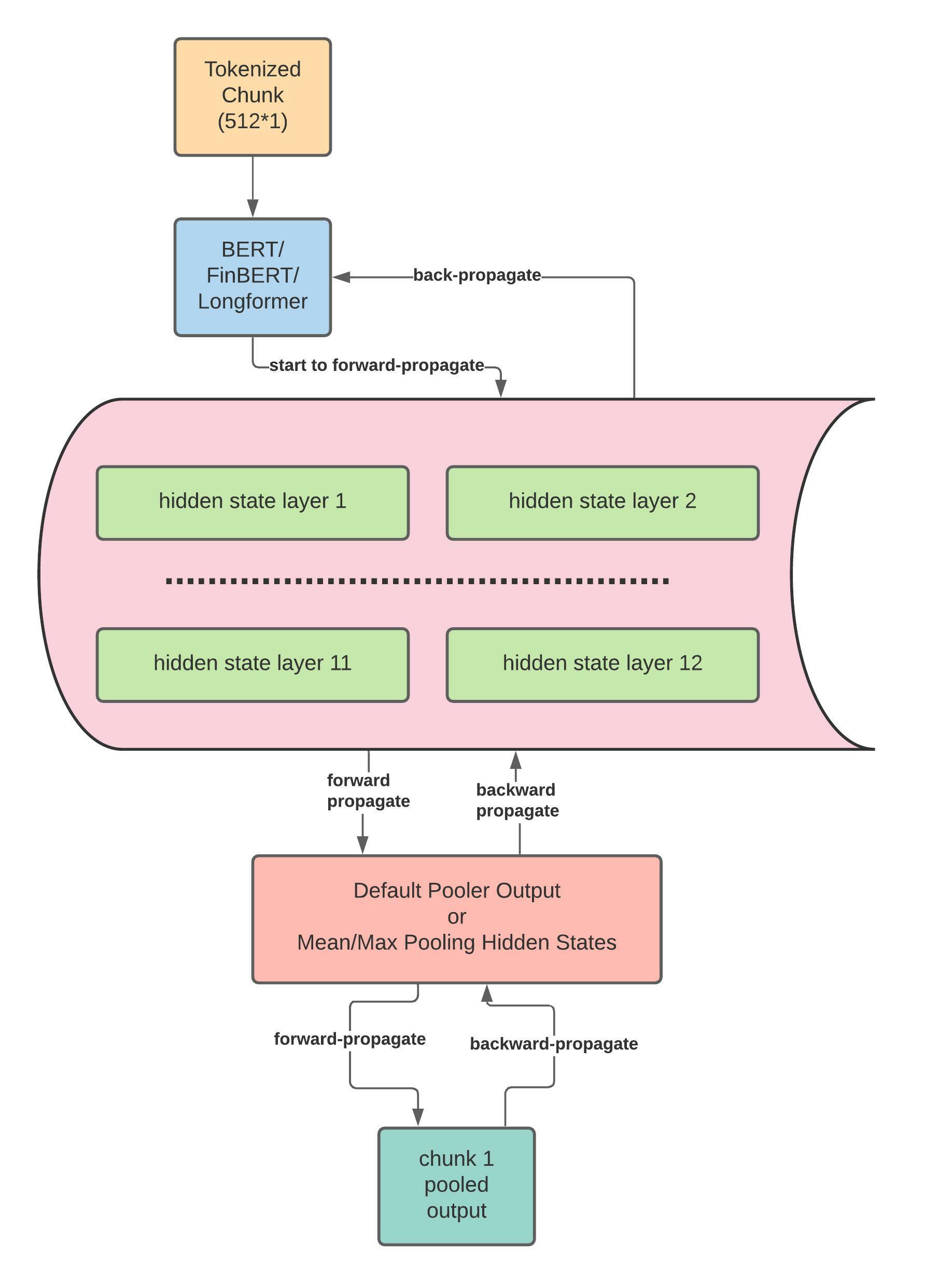}
  \caption{Chunk Level Language Model Pooling For Chunk Embeddings.}
  \label{fig:pooling}
\end{figure}

Given the sequence of chunks for a document, $\{c_1, c_2, \cdots, c_{m_i}\}$,
we need to convert these into features vectors $\{\mathbf{c}_1,
\mathbf{c}_2, \cdots, \mathbf{c}_{m_i}\}$, that represent the token sequence in
each respective chunk as a whole. We use SOTA language models like
BERT~\cite{BERT}, Longformer~\cite{Longformer}, and FinBERT~\cite{Yang} to
generate contextual token and chunk embeddings. We thus input each chunk
into the underlying language model, which typically outputs 12 hidden state
layers $\{l_1, l_2, \cdots, l_{12}\}$, where $l_i$ denotes layer $i$. The
output of each of these layers contains $b+2$ hidden state vectors
$\{\mathbf{z}_0^l, \mathbf{z}_1^l, \cdots, \mathbf{z}_{b+1}^{l}\}$, for $b+2$
tokens in the chunk, each of which has a size of 768, which is the
dimensionality of the hidden states. The language model also yields a
default pooler output, which is the embedding vector for the \texttt{<CLS>}
token, the first token, of the last hidden state layer after processing and
activation, denoted by $\mathbf{z}_0^{12}$. Figure~\ref{fig:pooling} shows
the schematic of how we use the underlying language model to generate the
hidden state layers, as well as the default pooler output, which are then
combined using various strategies outlined below to yield the chunk
embedding vector $\mathbf{c}_i$ for each chunk $c_i$ within each document.

Creating contextual embeddings is challenging, since a word can have
different meanings in different contexts. So it is important to first create
contextual token embeddings and then experiment with different strategies to
generate different chunk representations from these contextual embeddings.
We therefore studied several approaches for creating the final chunk
embedding vectors $\mathbf{c}_i$: 

\begin{itemize} 

\item {\it Default pooler output:} Since the \texttt{<CLS>} token embedding
    is an attention-weighted aggregation of all the tokens in a given chunk,
    each chunk $c_i$ can therefore be represented by the default pooling
    output vector $\mathbf{z}_0^{12}$ as the chunk embedding vector
    $\mathbf{c}_i$. The size of $\mathbf{c}_i$ is equal to the default
    hidden layer size of 768.

\item {\em Pooled hidden layers}: The empirical evaluation conducted in
    BERT-as-a-Service \cite{xiao2018bertservice} shows that using the last
    hidden layer gives the highest accuracy, but they also observed that it
    could also be more biased since it is the closest layer to the output
    layer. Hence, it is advisable to select the second-to-last hidden layer
    or a combination of different layers. In implementing this idea in
    practice, we take the set of all $b+2$ hidden state vectors from the
    penultimate layer, namely, $\{\mathbf{z}_0^{11}, \mathbf{z}_1^{11}, \cdots,
    \mathbf{z}_{b+1}^{11}\}$ and mean/max pool them into one vector of size
    768, which, after some non-linear activation, can be used as the chunk
    embedding vector $\mathbf{c}_i$. In addition, we can also follow a
    similar approach by selecting the last four hidden layers, namely
    $\{\mathbf{z}_0^{9}, \mathbf{z}_1^{9}, \cdots, \mathbf{z}_{b+1}^{9}\}$,
    $\{\mathbf{z}_0^{10}, \mathbf{z}_1^{10}, \cdots, \mathbf{z}_{b+1}^{10}\}$,
    $\{\mathbf{z}_0^{11}, \mathbf{z}_1^{11}, \cdots,\mathbf{z}_{b+1}^{11}\}$,
    and $\{\mathbf{z}_0^{12}, \mathbf{z}_1^{12}, \cdots,
    \mathbf{z}_{b+1}^{12}\}$, and produce four mean/max pooled vectors in
    the same way. These four vectors and mean/max are pooled into one
    vector, which on activation can be used as the chunk embedding vector
    $\mathbf{c}_i$. 
\end{itemize}


\subsection{Document-Level Attention Pooling}
\begin{figure}[!ht]
\centering
  \includegraphics[scale = 0.7]{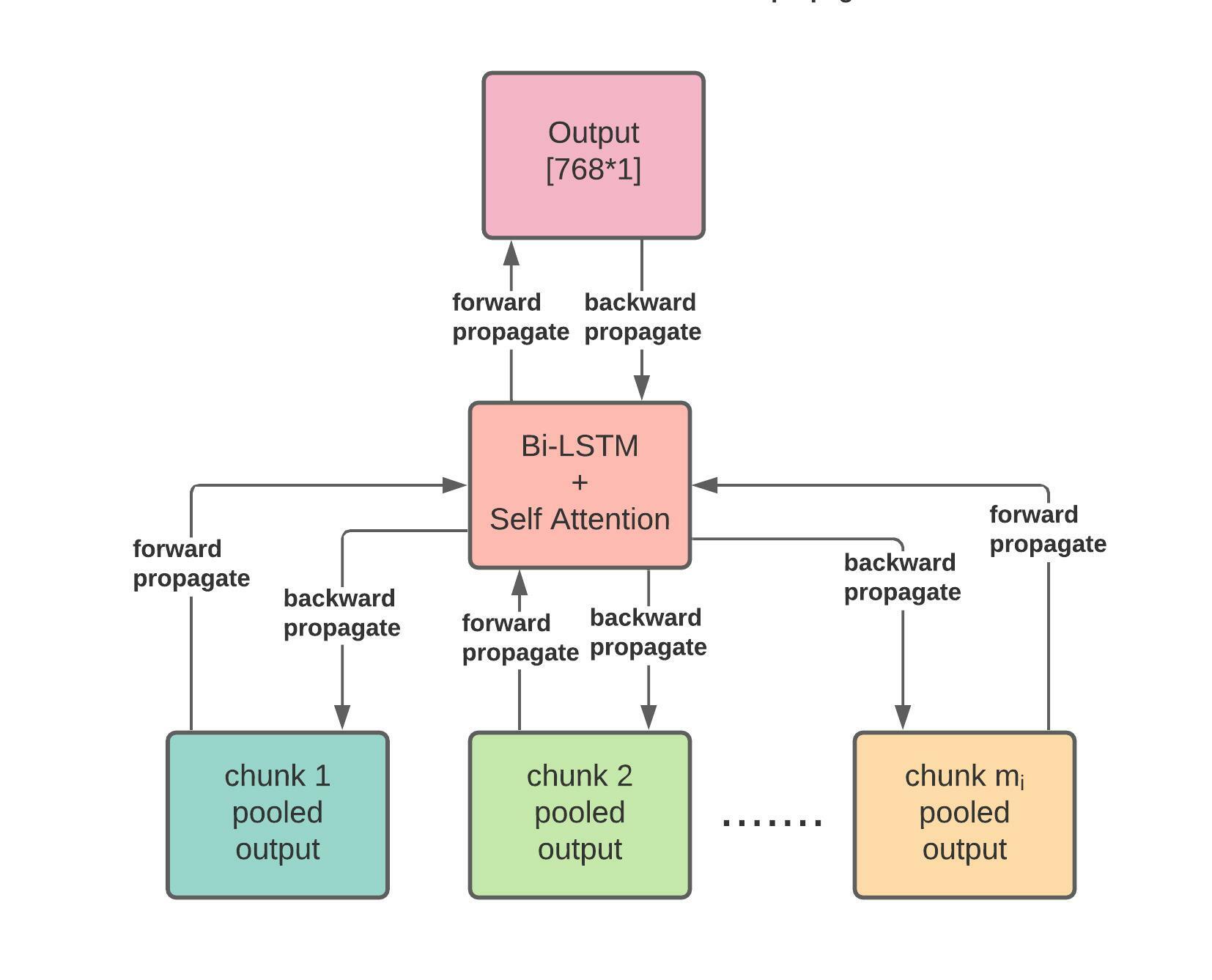}
  \caption{Attention Pooling via Bi-LSTM For Document Embeddings.}
  \label{fig:bilstm}
\end{figure}

Given the chunk embedding vectors $\{\mathbf{c}_1, \mathbf{c}_2, \cdots,
\mathbf{c}_{m_i}\}$, we need to aggregate them into an effective document
vector $\mathbf{d}_i$ for document $d_i$. Since the chunks are sequential in
nature, we can accomplish this using a recurrent Bi-LSTM model. However, not
all chunks in a long document are equally important. It is crucial to score
the chunks based on their importance in the document. For this, we introduce
chunk-level attention within the Bi-LSTM model. Given a document, we input
its chunk feature vectors $\{\mathbf{c}_1, \mathbf{c}_{2}, \cdots,
\mathbf{c}_{m_i}\}$ into the Bi-LSTM model. The output and hidden state
vectors of the Bi-LSTM for chunk $i$ are then obtained by concatenating the
outputs and the hidden states in forward and backward pass, respectively.
Formally, 

$\mathbf{o}_{i} = \overrightarrow{\mathbf{o}_{i}} 
\oplus \overleftarrow{\mathbf{o}_{i}}
\mathbf{h}_{i} = \overrightarrow{\mathbf{h}_{i}} \oplus \overleftarrow{\mathbf{h}_{i}}$
where $\oplus$ denotes concatenation, $\overrightarrow{}$ denotes forward
and $\overleftarrow{}$ denotes backward models, and the $\mathbf{h}_i$ and
$\mathbf{o}_i$ denote the hidden and output state vectors for chunk $c_i$,
respectively ($i$ also denotes the $i$-th element of the chunk sequence).
The attention score ${\alpha}_{i}$ for each chunk is calculated by taking
softmax over the product of outputs with the hidden state context vector.
The document feature vector $\mathbf{d}_{i}$ (of size 768) is obtained by
taking the weighted sum of the chunks according to their attention scores,
normalized by the number of chunks for that document. Formally,

${\alpha}_{i}  = \texttt{softmax}\Bigl(
\bigl\{\mathbf{o}_{1}^T\mathbf{h}_{i},
\mathbf{o}_{2}^T\mathbf{h}_{i},
\cdots,
\mathbf{o}_{{m_i}}^T\mathbf{h}_{i}
\bigr\}
\Bigr)\\
\mathbf{d}_{i}  = 
\frac{\sum_{j=1}^{{m}_{i}} 
{\alpha}_{j} \cdot \mathbf{c}_{j}}{{m}_{i}}$

Figure~\ref{fig:bilstm} shows an illustration of the document level
attention pooling step. At the bottom are the chunk embedding vectors
$\mathbf{c}_i$ as inputs, which are passed to the Bi-LSTM and attention
modules to create the document embedding $\mathbf{d}_i$.


\subsection{Model Training}\label{sec:3.4}

In the final stage of training, 
we feed each 768-dimensional document feature vector $\mathbf{d}_{i}$ to two
additional fully connected linear layers ${F\!C}_1$ and ${F\!C}_2$ (see
Figure~\ref{fig:overall}), with size 601
and 1, respectively, with a leaky ReLU
activation and a dropout layer applied to ${F\!C}_1$. The last layer ${F\!C}_2$
represents the output neuron to predict a target numeric variable. 
In other words, we concatenate
the historic score ${y}^{hist}$ (e.g., the previous year's value for stock
volatility or return on assets, etc.) with the document  vector
$\mathbf{d}_{i}$ so as to use both the numerical and textual features.
Formally,
\begin{equation}
    \mathbf{d}_i= \mathbf{o}_{{F\!C}_1} \oplus y\textsuperscript{\emph{hist}}
\end{equation}
where $\mathbf{o}_{{F\!C}_1}$ denotes the output features vector from ${F\!C}_1$.
Hence, ${F\!C}_1$ has 601 neurons, the first 600 of which are textual features,
and the last one is the historical numeric value, all of 
which are input to ${F\!C}_2$ to predict the target numeric
score $\hat{y}$. The loss function is MSE or mean squared error between the
predicted and true target value.




\section{FETILDA: Experiments}\label{sec4}

We now showcase the effectiveness of our FETILDA framework on text
regression tasks on very long financial documents. All of our experiments
were conducted on a machine with 2.5Ghz Intel Xeon Gold 6248 CPU, 768GB
memory, and a NVIDIA Tesla V100 GPU with 32GB memory. The neural network
models are implemented using PyTorch  v1.10 (\url{pytorch.org}) and the
HuggingFace library (\url{huggingface.co}) (for the BERT and Longformer
language models). Our code and datasets are publicly available on github via
\url{https://github.com/Namir0806/FETILDA}.

\subsection{Data Description: 10-K Reports}
A 10-K is a comprehensive report filed annually by a publicly traded company
about its financial performance and is required by the U.S. Securities and
Exchange Commission (SEC). \cite{kenton_2021} The SEC requires this report
to keep investors aware of a company's financial condition and to allow them
to have enough information before they buy or sell shares in the
corporation, or before investing in the firm’s corporate bonds. 
Generally,
the 10-K includes five sections \cite{kenton_2021}:

\begin{itemize}

\item \textbf{Business: }This provides an overview of the company’s main
    operations, including its products and services (i.e., how it makes
    money).

\item \textbf{Risk factors: }These outline any and all risks the company
    faces or may face in the future. The risks are typically listed in order
    of importance.
    
\item \textbf{Selected financial data: }This section details specific
    financial information about the company over the last five years. This
    section presents more of a near-term view of the company’s recent
    performance.
    
\item \textbf{Management’s discussion and analysis} of financial condition
    and results of operations: Also known as MD\&A, this gives the company
    an opportunity to explain its business results from the previous fiscal
    year. This section is where the company can tell its story in its own
    words.
    
\item \textbf{Financial statements and supplementary data: }This includes
    the company’s audited financial statements including the income
    statement, balance sheets, and statement of cash flows. A letter from
    the company’s independent auditor certifying the scope of their review
    is also included in this section.

\end{itemize}

The government requires companies to publish 10-K forms so investors have
fundamental information about companies so they can make informed investment
decisions. This form gives a clearer picture of everything a company does
and what kinds of risks it faces \cite{kenton_2021}. However, the length of
10-K reports has generally increased dramatically in recent years. According
to a Wall Street Journal article \cite{monga_chasan_2015}, the average 10-K
report is getting longer, from about 30,000 words in 2000 to about 42,000
words in 2013. In the article, GE finance chief Jeffrey Bornstein is
reported to have said that not a retail investor on planet earth could get
through it, let alone understand it.

Therefore, our goal is to extract the soft information contained in the
textual data of these extremely lengthy 10-K reports, in order to better our
predictions of forward-looking KPIs. Fortunately, not all sections of the
10-K report are useful in predicting the future performance of banks and
companies using the textual features they contain. The business section does
not really give much information on expected future performance. Both the
selected financial data section, and the financial statements and
supplementary data section contain information about expected future
performance of the company, but they mainly contain quantitative data in
numeric form.

However, the risk factor section (Item 1/1A) and the management's discussion
and analysis (MD\&A) section (Item 7) of the 10-K report, which are the
sections we utilize in this paper, are worthy of notice, and can very well
contain a treasure trove of soft information that humans are not able to
understand in a quantifiable way. Item 1/1A contains the company's view of
the various risk factors that can impact the company in the future, and Item
7 contains the company's view of how it has performed in the past year and
how it expects to perform in the coming year. These two sections
usually contain forward-looking statements, and the contexts of these two
sections -- the way things are said, how views are articulated -- are
also indicative of what the company thinks of itself internally. The
experiments done by Tsai and Wang \cite{tsai2017risk} have shown that using
only the MD\&A section (Item 7) produces equivalent results, compared to
using the whole 10-K document, and for comparison, we replicate their
experiment with our approach below \ref{subsect:compresults}. This is why we
focus on these two sections of the 10-K report. Since the industry standard
is to only use quantitative data to predict future KPIs, we want to add the
qualitative data coming from text into the mix, in order to achieve better
predictions.


\subsection{Datasets and Target Metrics}
\label{sec:dataset}

\subsubsection{US Banks Dataset}
We collected the 10-K filings for all US banks for the period between 2006
and 2016 (from the SEC EDGAR website: \url{www.sec.gov/edgar}), as well as
the corresponding quantitative target data from the WRDS Center for Research
in Security Prices \url{wrds-www.wharton.upenn.edu}. 
For US banks our goal is to predict several KPI metrics
using the 10-K reports. In particular, we focus on eight metrics that
indicate either the performance or risk of a given bank: Return on Assets
(ROA), Earnings per Share (EPS), Return on Equity (ROE), Tobin's Q Ratio,
Tier 1 Capital Ratio, Leverage Ratio, Z-Score, and Market-to-Book Ratio. The
target metrics are defined below.

\begin{itemize}
    \item \textbf{Return on Assets (ROA): }
    ROA is calculated by dividing a company’s net income by total assets:
    \begin{equation}
    ROA =\frac{Net \, Income}{Total \, Assets}
    \end{equation}
    Higher ROA shows more asset efficiency and productivity. ROA varies and
    is highly dependent on the industry. Thus, it is best to compare ROA
    with a company's previous ROA or with a similar company's ROA. It has a
    limitation that it cannot be used across industries. For instance,
    companies in the technology industry and oil drillers industry will have
    different asset bases.
    
    \item \textbf{Return on Equity (ROE):}
    According to \cite{ROE}, ROE is a measure of financial performance
    calculated by dividing net income by shareholders' equity:
    \begin{equation}
    ROE = \frac {Net \,Income} {Total \,Equity} 
    \end{equation}
    Since equity is simply the assets of a company minus the debt, ROE is
    basically the return on investment to the shareholders of a company. ROE
    is an indicator of a company's profitability and efficiency in
    generating its profits.
    
    \item \textbf{Earning per share (EPS):}
    EPS is an indicator of a company's profitability. It is calculated as a
    company's profit divided by the outstanding shares of its common stock: 
    \begin{equation}
    EPS =\frac{Net \,Income - Preferred \,Dividends }{End\text{-of}\text{-Period} \,Common \,Shares \,Outstanding }
    \end{equation}
    The higher a company's EPS, the more profitable per share it is.
    
    \item \textbf{Tobin's Q Ratio (TQR):}
    TQR represents the ratio of the market value of a firm's assets to the
    replacement cost of the firm's assets:
    \begin{equation}
    Tobin's \,Q \,Ratio =\frac{
    Equity \,Market \,Value+ Liabilities\, Book\, Value}{Equity\, Book\, Value+Liabilities\, Book\, Value}
    \end{equation}
    This ratio indicates how the market views the managers' prospects of
    using firm's asset to generate future value for investors of the firm.

    \item \textbf{Leverage Ratio (LR):}
    The Leverage Ratio measures the extent of debt financing for a firm,
    therefore assesses the ability of a company to meet its financial
    obligations. The potential downside of debt financing is that more debt
    poses a threat to a firm's viability if its earnings can't support the
    dues on the debt. The leverage ratio measures the extent of debt
    financing used by a firm:    
    \begin{equation}
    Leverage \,Ratio =\frac{Average \,Total\, Assets}{Average\, Equity }
    \end{equation}
    A high ratio means the firm is using a large amount of debt to finance
    its assets and a low ratio means the opposite. The most leveraged
    institutions in the US include banks. There is a limit on a bank's
    lending capacity. Three different regulatory bodies, the Federal Deposit
    Insurance Corporation (FDIC), the Federal Reserve, and the Comptroller
    of the Currency, review and restrict the leverage ratios for US banks.
    They restrict the bank's lending compared to how much capital the bank
    assigns to its own assets. This is important because banks can ``write
    down'' the capital part of their assets if there is a drop in total
    asset value. As such, assets financed by debt cannot be written down
    since these funds are owed to the bank's bondholders and depositors. The
    guidelines for bank holding companies created by the Federal Reserve are
    complicated and vary depending on the bank's rating. These guidelines
    have become stricter since the Great Recession of 2007-2009 when banks
    that were ``too big to fail'' necessitated banks to be made more
    solvent. Such restrictions limit the bank's lending because it is more
    difficult and expensive for a bank in raising capital than borrowing
    funds.
    
    \item \textbf{Tier 1 Capital Ratio (T1CR):}
    The Tier 1 Capital Ratio is a measure of a bank's financial strength
    from a regulator's point of view that was adopted as part of the Basel
    III Accord on bank regulation. The 2007-2009 crisis showed that many
    banks had insufficient capital to take in the losses or remain liquid,
    and were backed by too much debt and not enough equity. Thus the Basel
    III standard was enforced so as to increase bank's capital buffers and
    make sure that they are able to withstand financial distress before
    becoming insolvent. Tier 1 capital is bank's equity capital and
    disclosed reserves. The Tier 1 capital ratio is the ratio of a bank’s
    core Tier 1 capital to its total risk-weighted assets:
      \begin{equation}
    Tier \,1 \,Capital \,Ratio =\frac{Tier\,1 \,Capital}{Total \,Risk-Weighted \,Assets}
    \end{equation}
    These risk-weighted assets include all assets that are systematically
    weighted for credit risk.
 
    \item \textbf{Z-score (Z):}
    The Z-score links a bank's capitalization with its return ($ROA$) and
    risk (volatility of returns). Z-score \footnote{This Z-score should not
    be confused with the Altman Z-score \cite{altman1968financial}. The
Altman Z-score is a set of financial and economic ratios and it is mainly
used as a predictor of corporate finance distress.} is a popular indicator
of bank risk and it was proposed in \cite{roy1952safety}. The basic idea of
the Z-score is to relate a bank’s capital level to variability in its
returns, in order to know how much variability in returns a bank can absorb
without becoming insolvent. This variability in returns is typically
measured by the standard deviation of Return on Assets ($ROA$). As per its
definition, insolvency occurs when the firm's losses exceed its equity
$-L>E$, where $-L$ is loss and $E$ is equity. The Z-score looks at Return on
Assets ($ROA$) and capital-to-assets ratio ($CAR=E/Assets$) to measure the
overall bank risk: 
    \begin{equation}
    Z\text{-Score}= \frac{ROA + CAR}{\sigma(ROA)} 
    \end{equation}
    where, $\sigma(ROA)$ is the standard deviation of $ROA$ for a specific time period.

    \item \textbf{Market-to-Book Ratio (MBR):}
    The Market-to-Book Ratio is used to evaluate a company’s current market
    value relative to its book value, and is calculated by dividing the
    current stock price of all outstanding shares (i.e., the price that the
    market believes the company is worth) by the book value:
    \begin{equation}
    Market\text{-to}\text{-Book}\, Ratio = \frac {Market \,Capitalization} {Total \,Book\, Value} 
    \end{equation}
    The Market-to-Book Ratio shows the financial valuation of a company's
    stock, and is an indicator of how much equity investors value each share
    relative to their book value.
\end{itemize}

While the entire 10-K report is a very long disclosure document, as noted
above, Items 1A and 7/7A are considered as important subsections in a
10-K report \cite{amel2016information}. Item 1A (Risk Factors) 
includes information about the most significant risks for a company or
its securities. The risk factors are typically reported in order of their
importance. However, it focuses on the risks themselves, and not necessarily
on how the company addresses those risks. Some risks apply to the entire
economy, some only to the specific industry sector or region, and some are
directly related to the company. Item 7 (MD\&A) 
gives the company's perspective on the business results of the past
financial year. The MD\&A subsection is meant for the management to relate
in its own words the analysis of their financial condition. Finally, Item 7A
(Quantitative and Qualitative Disclosures about Market Risk)
provides information about the company's exposure to market risk, such as
interest rate risk, foreign currency exchange risk, commodity price risk or
equity price risk. These subsections are themselves also quite long. The
dataset statistics for the 10-K reports for all US Banks for the period of
2006-2016 are reported in Table \ref{tab:doc-stats}.

\begin{table}[!ht]
\caption{US bank dataset statistics.}
\label{tab:doc-stats}
\centering
\begin{tabular}{|l|c|c|}
\hline
                                                                           & Item 1A & Item 7/7A \\ \hline
number of total documents                                                & \multicolumn{2}{c|}{5321}       \\ \hline
after extracting items                                            & \multicolumn{2}{c|}{3396}       \\ \hline
target data available                                        & 2479           & 2500           \\ \hline
\begin{tabular}[c]{@{}l@{}}average document length\end{tabular} & 4435.69        & 12589.75       \\ \hline
\end{tabular}

\end{table}
\begin{figure}[!ht]
    \centering
    \includegraphics[width=3in, height=2.3in]{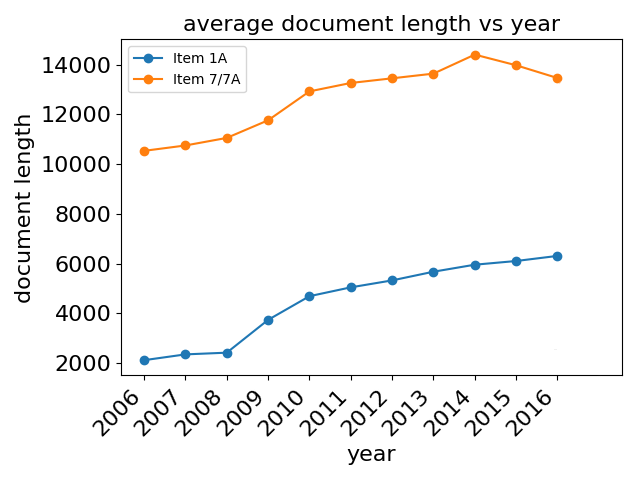}
    \vspace{-0.2in}
    \caption{The average document length (number of words) increases with time for both Item 1A and Item 7/7A.}
    \label{fig:doc-len}
\end{figure}

The 10-K reports for US Banks (2006-2016) total 5321 documents, but not all
reports have both the Item 1A or 7/7A subsections. Out of the total, only
3396 10-K reports have both these important subsections. Furthermore, we
found that not all banks have all the eight target KPI values that we need
for regression. Out of the 3396 documents, we have 2479 Item 1A and 2500
Item 7/7A with their eight metrics in full as target data, which makes up
the final document set used in our experiments. The average document length
(in terms of the number of words) is 4436 for Item 1A and 12590 for Item
7/7A, as noted in Table \ref{tab:doc-stats}. Furthermore, Figure
\ref{fig:doc-len} shows how the average document length increases in time.
We sort the documents chronologically from 2006 to 2016, 
and choose the first 80\% of the data for training, and the remaining 20\%
as validation and testing data, with a 50/50 split between the latter two.
In terms of target data normalization, for each and every one of the eight
target metrics, we performed min-max scaling to normalize the data for
training.

\subsubsection{FIN10K Dataset \cite{FIN10K}}

In addition to our dataset of 10-K reports filed by US banks, we also
compare our FETILDA approach on the regression task outlined in Tsai and
Wang \cite{tsai2017risk} using the dataset in the FIN10K project
\cite{FIN10K}, which contains Item 7 of 10-K reports of US companies from
1996 to 2013 and the stock return volatilities twelve months before and
after each report.
Table \ref{tab:FIN10Kstats} shows the statistics for the part of the FIN10K
dataset \cite{FIN10K} used in our comparative experiments against the LOG1P+
approach taken in Tsai and Wang  \cite{tsai2017risk}.

\begin{table}[!ht]
\centering
\caption{FIN10K dataset \cite{FIN10K} statistics.}
\label{tab:FIN10Kstats}
\centering
\resizebox{\linewidth}{!}{%
\begin{tabular}{|l|r|r|r|r|r|r|r|r|}
\hline
&1996 - 2000 &2001 &2002 &2003 &2004 &2005 &2006 \\
\hline
number of total documents &8703 &1825 &2023 &2866 &2861 &2698 &2564 \\
\hline
average document length &5079.4 &6245.6 &8414.3 &10324.7 &11499.6 &12528.1 &12198.1 \\
\hline
\end{tabular}}
\end{table}

\begin{figure}[!ht]
    \centering
    \includegraphics[width=3in, height=2in]{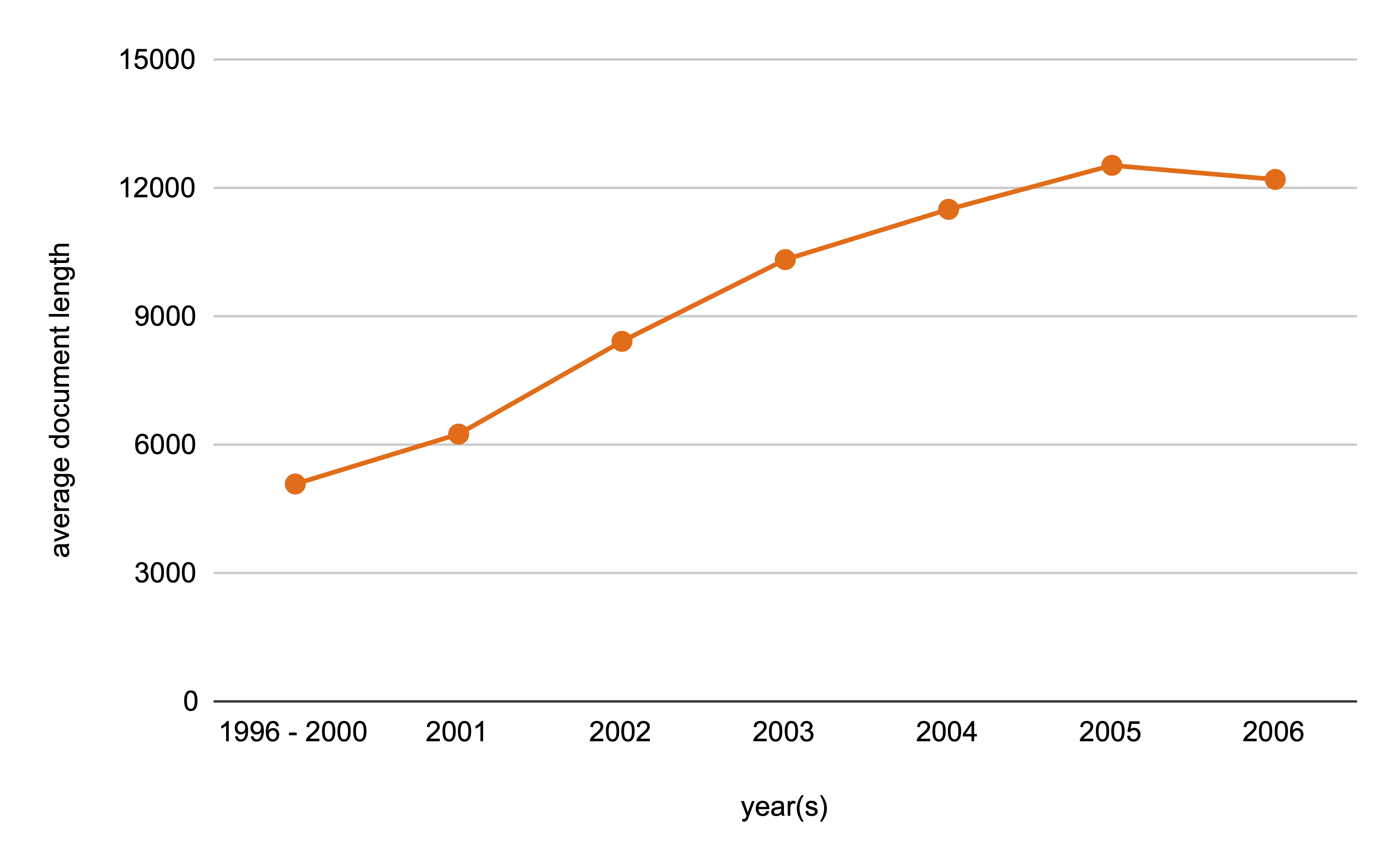}
    \vspace{-0.2in}
    \caption{The average document length (number of words) increases with time for Item 7 of 10-K reports in FIN10K \cite{FIN10K}.}
    \label{fig:fin10k-doc-len}
\end{figure}

To replicate the regression results in \cite{tsai2017risk}, we follow
the same experimental setup, and therefore use the reports from 1996 to
2000 as training and validation data, and reports for each year from 2001 to
2006 as separate testing data. In addition, we did not perform any target
data normalization, in order to replicate the experiment completely.
Further, we choose the first 80\% of reports from 1996 to 2000 as training
data, and the remaining 20\% as validation data. As we can see, the number
of documents in the training and validation data from 1996 to 2000 is more
than three times as many as that of the US banks dataset, but the average
document length is significantly smaller than that of Item 7/7A in the US
banks dataset. In the testing data, from 2001 to 2006, the number of
documents is generally increasing, as well as the average document length,
as shown in Figure \ref{fig:fin10k-doc-len}.

In terms of the regression task, Tsai and Wang \cite{tsai2017risk}
experimented on stock return volatilities twelve months before and after
each report, and we use the same target values in our experiment. According
to Tsai and Wang, volatility is a common risk metric defined as the standard
deviation of a stock's returns over a period of time. Historical
volatilities are derived from time series of past stock market prices as a
proxy for financial risk. Let $S_t$ be the price of a stock at time $t$.
Holding the stock for one period from time $t - 1$ to time $t$ results in a
simple net return of $R_t = \frac{S_t}{S_{t-1}}-1$ \cite{tsay2005analysis}.
Therefore, the volatility of returns for a stock from time $t - n$ to $t$ is
defined as
\begin{equation}
    v_{[t-n,t]} = \sqrt{\frac{\sum_{i=t-n}^{t}(R_i - \bar{R})^2}{n}}
\end{equation}
where $\bar{R} = \sum_{i=t-n}^{t}R_i/(n+1)$. 

\subsection{Methods}
We now outline the results of our framework on both the US banks and FIN10K
datasets. Overall, we experiment with seven different methods, the first
three being baseline methods with which we compare the last four to evaluate
the performance of our approach. In order to effectively compare different
methods, all results report the mean squared error (MSE).  The methods are
as follows:

\begin{itemize}
    \item {\bf TF-IDF}~\cite{Jurafsky21}: In this baseline method, we use
        the term frequency - inverse document frequency features, which help
        in scoring important words, concatenated with the historical
        score ${y}^{hist}$, and then apply regression on the features
        to predict the target values. There are three regression methods we
        experiment with: (1) Linear Regression, (2) Support Vector
        Regression and (3) Kernel Ridge Regression. We use validation data
        to choose the best regression method among the three, according to
        which one achieves the lowest validation loss.
    
    \item {\bf Linear Regression}~\cite{zaki2020data}: In this baseline
        method, we simply take the historical score ${y}^{hist}$ and run
        (bivariate) linear regression on it to predict the target variable.
        This method therefore utilizes only numerical data.
    
    \item {\bf LOG1P+}~\cite{tsai2017risk}: This is the method used in the
        volatility regression task proposed by Tsai and Wang
        \cite{tsai2017risk}. The word features are formed using LOG1P,
        calculated as $LOG1P = log(1 + TC(t,\mathbf{d}))$, where
        $TC(t,\mathbf{d})$ denotes the term count of a word $t$ in a given
        document $\mathbf{d}$. Furthermore, the logarithm of the stock
        return volatility twelve months before each report is used as an
        additional numeric feature, and together, the word features and
        numeric features are input into a Support Vector Regression model.
    
    \item {\bf FETILDA w/ BERT}: In this method, we use our approach,
        detailed in section \ref{sec3}, with plain BERT \cite{BERT} as the
        underlying language model, setting the chunk size to 512 tokens and
        using the default pooling method.
    
    \item {\bf FETILDA w/ FinBERT}: Here we used our approach with FinBERT
        \cite{Yang} as the LM, which was pre-trained on 10-K, 10-Q, and
        analyst reports, setting the chunk size to 512 tokens and using the
        default pooling method.
    
    \item {\bf FETILDA w/ Longformer}: Now, to test the effectiveness of a
        bigger block size with a pretrained model, we use our approach with
        Longformer \cite{Longformer} as the underlying language model,
        setting the chunk size to 4096 tokens and using the default pooling
        method.
    
    \item {\bf FETILDA w/ Nystromformer}:
        Finally, to test the effectiveness of an even bigger block size, but
        without a pretrained model (that is, training from scratch), we use
        our approach with Nystromformer \cite{xiong2021nystromformer} as the
        underlying language model, setting the chunk size to 8192 tokens,
        the number of layers to one, the number of attention heads to eight,
        and using the default pooling method.
\end{itemize}

With all four versions of FETILDA, namely using BERT, FinBERT, Longformer,
and Nystromformer, we performed an extensive set of experiments, evaluating
our approach in predicting all eight different KPI metrics for the US banks
dataset, and stock return volatility for the FIN10K dataset \cite{FIN10K}.
For the US banks dataset, the historical scores
are numeric values of each of the eight metrics in the previous year of the
report, and for the FIN10K dataset, they are the stock return volatilities
twelve months before each report. 
In addition to applying our approach as described in subsection
\ref{sec:3.2} with fully unfrozen LM layers, enabling model fine-tuning, we
also report the effect of freezing all the LM layers and freezing only the
last layer in FETILDA when we apply it on the US banks dataset, with both
Item 1A and Item 7/7A. This allows us to compare the effect of fine-tuning versus
the default pretraining approach.

\subsection{Comparative Performance Results}\label{subsect:compresults}
In all four versions of FETILDA, we train the model with eight varying
learning rates from 0.0006 to 0.0013, and four different final layer options
detailed in subsection \ref{sec:3.4}, and pick the epoch and parameters with
the best validation loss. Due to the memory constraint of 32 GB, for a given
document, the GPU can only process up to around 20,480 tokens at a time, so
we truncate the rest if a document goes beyond that length. However, this
only happens for a minority of cases in our experiments, and we do not
truncate at all in our experiments with fully frozen language models. As
mentioned above, we use the default pooling strategy to extract chunk
embedding vectors, and then use the Yang et al. FinBERT~\cite{Yang} model.
We empirically show below that both these choices are in fact the best ones
among the different pooling and FinBERT variants, respectively. Finally, for
both FETILDA (w/BERT, w/FinBERT, w/LongFormer, and w/Nystromformer) and
TF-IDF/LOG1P+ we select the best among the following regression models based
on the validation data: (1) Linear Regression, (2) Support Vector
Regression, using a RBF Kernel with $C=0.1$ and $\epsilon=0.0001$, and (3)
Kernel Ridge Regression, using a RBF Kernel with $\alpha=0.1$ and
$\gamma=0.1$. For FETILDA, we also include the variant based on the
predicted output (from ${F\!C}_2$) with MSE loss.


\begin{table}[!ht]
\centering
\caption{MSE results on Item 7/7A. Best results in bold per KPI.}
\label{tab:sec7hist}
\small
\resizebox{\linewidth}{!}{%
\begin{tabular}{|l|r|r|r|r|r|r|r|r|r|}

\hline
Models\textbackslash Metrics &ROA &ROE &EPS &TQR &T1CR &LR &Z  &MBR \\
\hline
TF-IDF &0.000879 &0.010422 &0.001022 &0.022000 &0.000767 &0.002594 &0.028926 &0.005765 \\
LOG1P+ &0.001112 &0.025760 &0.001887 &0.026116 &0.006582 &0.005101 &0.033905 &0.020450 \\
Linear Regression &0.001432 &0.010096 &0.001564 &0.022587 &0.000306 &0.002441 &0.030760 &0.005757 \\
\hline
FETILDA w/BERT (Fully Unfrozen) &0.000796 &0.009227 &0.000897 &0.021409 &0.000325 &0.002502 &0.029505 &0.005651 \\
FETILDA w/FinBERT (Fully Unfrozen) &\textbf{0.000746} &0.008901 &0.000932 &0.019150 &0.000317 &0.002535 &0.029516 &0.005657 \\
FETILDA w/Longformer (Fully Unfrozen) &0.000813 &0.008507 &0.000858 &0.017358 &0.000296 &0.002467 &\textbf{0.028697} &0.005683 \\
\hline
FETILDA w/BERT (Fully Frozen) &0.000890 &0.010052 &0.001109 &0.022748 &0.000328 &0.002581 &0.028966 &0.005950 \\
FETILDA w/FinBERT (Fully Frozen) &0.001093 &0.009401 &0.001906 &0.021882 &0.000447 &0.002514 &0.030094 &0.005695 \\
FETILDA w/Longformer (Fully Frozen) &0.000801 &0.008501 &0.000876 &0.019053 &0.000308 &0.002436 &0.028965 &0.005957 \\
\hline
FETILDA w/BERT (Last Layer Frozen) &0.000850 &0.009903 &0.000960 &0.021728 &0.000306 &0.002469 &0.029203 &0.005798 \\
FETILDA w/FinBERT (Last Layer Frozen) &0.000844 &0.008543 &0.000988 &0.021425 &0.000304 &0.002445 &0.029637 &0.005678 \\
FETILDA w/Longformer (Last Layer Frozen) &0.000849 &\textbf{0.008356} &\textbf{0.000851} &\textbf{0.016436} &0.000291 &0.002419 &0.029011 &0.005481 \\
\hline
FETILDA w/Nystromformer &0.000815 &0.008989 &0.000869 &0.017554 &\textbf{0.000264} &\textbf{0.002417} &0.030462 &\textbf{0.004302} \\
\hline

\end{tabular}}
\end{table}

\paragraph{Item 7/7A (MD\&A Section)} 
Table
\ref{tab:sec7hist} shows the performance comparison between the four
versions of our approach on Item 7/7A and baseline methods: TF-IDF and
LOG1P+ for textual modeling with historic scores, and linear regression for
numerical modeling. We can see that TF-IDF features always do better than
LOG1P+.
{\em However, for all metrics, our method outperforms the baseline methods
(TF-IDF, LOG1P+ and linear regression), with FETILDA w/Longformer
\cite{Longformer} performing the best in a majority of cases and FETILDA
w/Nystromformer coming as a close second}. In addition, we also see a
significant edge in the performance of FinBERT in the prediction of ROA
target values. Overall, FETILDA w/BERT performs best or second best on three
metrics, FETILDA w/FinBERT on four metrics, and FETILDA w/Longformer on six
out of the eight metrics.

Table~\ref{tab:sec7hist} also shows the effect of freezing either the last
layer or all layers for all three underlying LMs. Freezing all the layers
means that we use the pre-trained embeddings. On the other hand, the fully
unfrozen language model layers comprise the fine-tuning based approach,
since the entire model is fine-tuned for each of the KPIs during training. 
As we can see, the token
embeddings from Longformer with the last layer frozen perform the best, and
even outperform the fine-tuned Longformer model (with unfrozen layers) on
six out of the eight metrics, but fully freezing Longformer is not as
beneficial. On the other hand, BERT benefits from fine-tuning for five
out of the eight metrics. This may be due to the much longer context blocks
used in Longformer, which is able to capture more contextual information and
does not need too much fine-tuning on the downstream tasks. FETILDA
w/FinBERT benefits from fine-tuning in all of the eight metrics, either with
all layers unfrozen or freezing only the last layer, which shows the
effectiveness of taking a domain-specific pretrained language model and then
fine-tuning it for a particular downstream task. Overall, FETILDA
w/Longformer offers the best or close to best results in predicting seven
out of the eight KPI targets for US Banks.


\begin{figure}[!ht]
    \centering
    \includegraphics[scale=0.035]{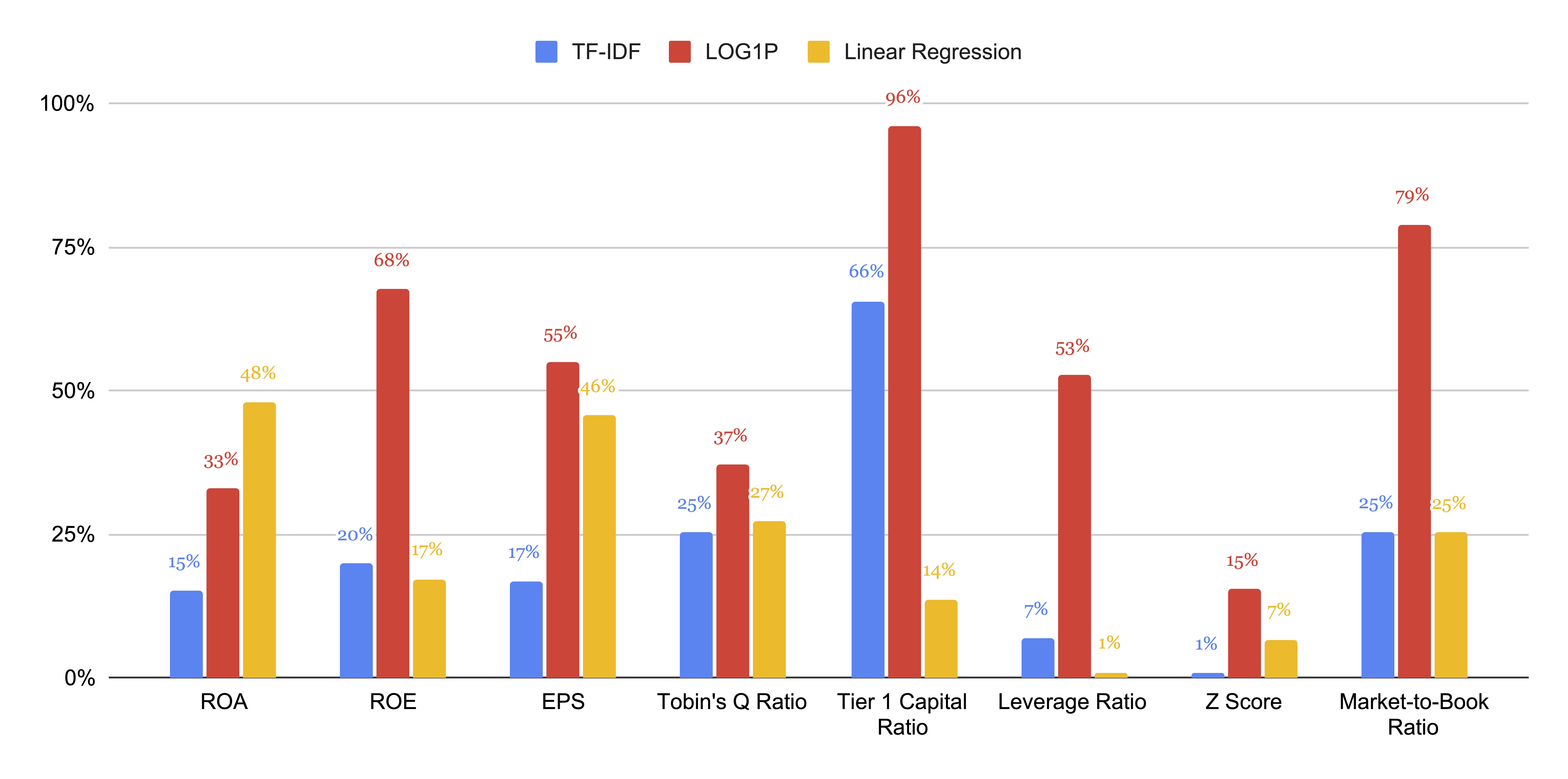}
    \caption{FETILDA Improvement on Item 7/7A over TF-IDF, LOG1P, and linear regression.}
    \label{fig:sec7histchart}
\end{figure}

To characterize the improvement due to FETILDA with different language
models, for every metric, we select the best performing version of FETILDA
in terms of mean squared error and calculate its improvement against the
text and numeric baseline methods using the formula:
$$(MSE_{baseline}-MSE_{FETILDA})/{MSE_{baseline}}$$ 
The results are shown in
Figure \ref{fig:sec7histchart}.
We observe large improvements for ROE, EPS, Leverage Ratio, and Tobin's Q
Ratio, using our approach over TF-IDF. For Tier 1 Capital ratio, our
approach achieves the highest improvement over TF-IDF in percentage terms.
Overall, our method outperforms TF-IDF, LOG1P+, and linear regression for
all metrics.

\paragraph{Item 1A (Risk Factors Section)} 
Next, we
report results on Item 1A. Table
\ref{tab:sec1Ahist} shows the performance comparison between the four
versions of our approach on Item 1A and the baseline methods, TF-IDF, LOG1P+
and linear regression. In six out of eight metrics, our method outperforms
both TF-IDF and linear regression, with Longformer \cite{Longformer}
performing the best in three cases and Nystromformer performing the best in
two cases. Overall FETILDA is best or second best in all eight cases.
The baseline TF-IDF method performs the best in two out of the eight tasks
with Item 1A, but is never the best for Item 7/7A. This may be due to the
difference in document lengths between Item 1A and Item 7/7A. Table
\ref{tab:doc-stats} shows that Item 7/7A is typically three times the length
of Item 1A. With a shorter document length, we posit that TF-IDF feature
vectors can represent Item 1A reasonably well, whereas the more complex
contextual language models such as BERT, FinBERT, or Longformer do not have
too much room for improvement. {\em Nevertheless, it is important to note
    that FETILDA w/Longformer is 
the best or second best in six out of the eight metrics}.

\begin{table}[!ht]
\centering
\caption{MSE results on Item 1A. Best results in
bold per KPI.}
\label{tab:sec1Ahist}
\small
\resizebox{\linewidth}{!}{%
\begin{tabular}{|l|r|r|r|r|r|r|r|r|r|}
\hline
Models\textbackslash Metrics &ROA &ROE &EPS &TQR &T1CR &LR &Z &MBR  \\
\hline
TF-IDF &\textbf{0.000770} &0.008785 &\textbf{0.000811} &0.016984 &0.000248 &0.002627 &0.030511 &0.005165 \\
LOG1P+ &0.001239 &0.024953 &0.001896 &0.026026 &0.005150 &0.005270 &0.035720 &0.021402 \\
Linear Regression &0.001407 &0.010174 &0.001577 &0.022500 &0.000299 &0.002534 &0.032102 &0.005802 \\
\hline
FETILDA w/BERT (Fully Unfrozen) &0.000811 &0.008520 &0.000820 &0.019151 &0.001353 &0.002559 &0.029614 &0.004944 \\
FETILDA w/FinBERT (Fully Unfrozen) &0.000867 &0.008671 &0.001171 &0.017383 &0.000385 &0.002560 &0.030583 &0.004937 \\
FETILDA w/Longformer (Fully Unfrozen) &0.000790 &0.007940 &0.000826 &\textbf{0.015620} &0.000937 &0.002527 &0.030130 &\textbf{0.004555} \\
\hline
FETILDA w/BERT (Fully Frozen) &0.000856 &0.008788 &0.001076 &0.018572 &0.000315 &0.010919 &0.030225 &0.004908 \\
FETILDA w/FinBERT (Fully Frozen) &0.000976 &0.008626 &0.001274 &0.018254 &0.000428 &\textbf{0.002471} &0.032155 &0.004911 \\
FETILDA w/Longformer (Fully Frozen) &0.000811 &0.008053 &0.000854 &0.018429 &0.000930 &0.002619 &0.034284 &0.004955 \\
\hline
FETILDA w/BERT (Last Layer Frozen) &0.000774 &0.007803 &0.000824 &0.017883 &0.000726 &0.002751 &0.029729 &0.004943 \\
FETILDA w/FinBERT (Last Layer Frozen) &0.000850 &0.008814 &0.000834 &0.018282 &0.000485 &0.002612 &0.030115 &0.004967 \\
FETILDA w/Longformer (Last Layer Frozen) &0.000795 &\textbf{0.007409} &0.000821 &0.018100 &0.000242 &0.002715 &0.030415 &0.004894 \\
\hline
FETILDA w/Nystromformer &0.000780 &0.007659 &0.000925 &0.016263 &\textbf{0.000226} &0.002831 &\textbf{0.029426} &0.005640 \\
\hline

\end{tabular}}
\end{table}

Freezing the layers of all
three underlying LMs produces different results for different LMs. None of
the frozen (pre-trained) embeddings from Longformer outperforms the
fine-tuned Longformer with unfrozen layers or only the last layer frozen in
all of eight metrics. The frozen (pre-trained) embeddings from BERT perform
better than its fine-tuned counterpart in two out of eight metrics, and
worse in the remaining six metrics, with more noticeable improvements in
predicting Tier 1 Capital Ratio target values. The frozen (pre-trained)
embeddings from FinBERT performs better than its fine-tuned counterpart in
three out of eight metrics, and worse for the remaining five metrics, with a
slightly significant improvement in predicting Leverage Ratio target values.
Overall, with frozen layers, FETILDA is best or second best in five out of
the eight metrics.


\begin{figure}[!ht]
    \centering
    \includegraphics[scale=0.035]{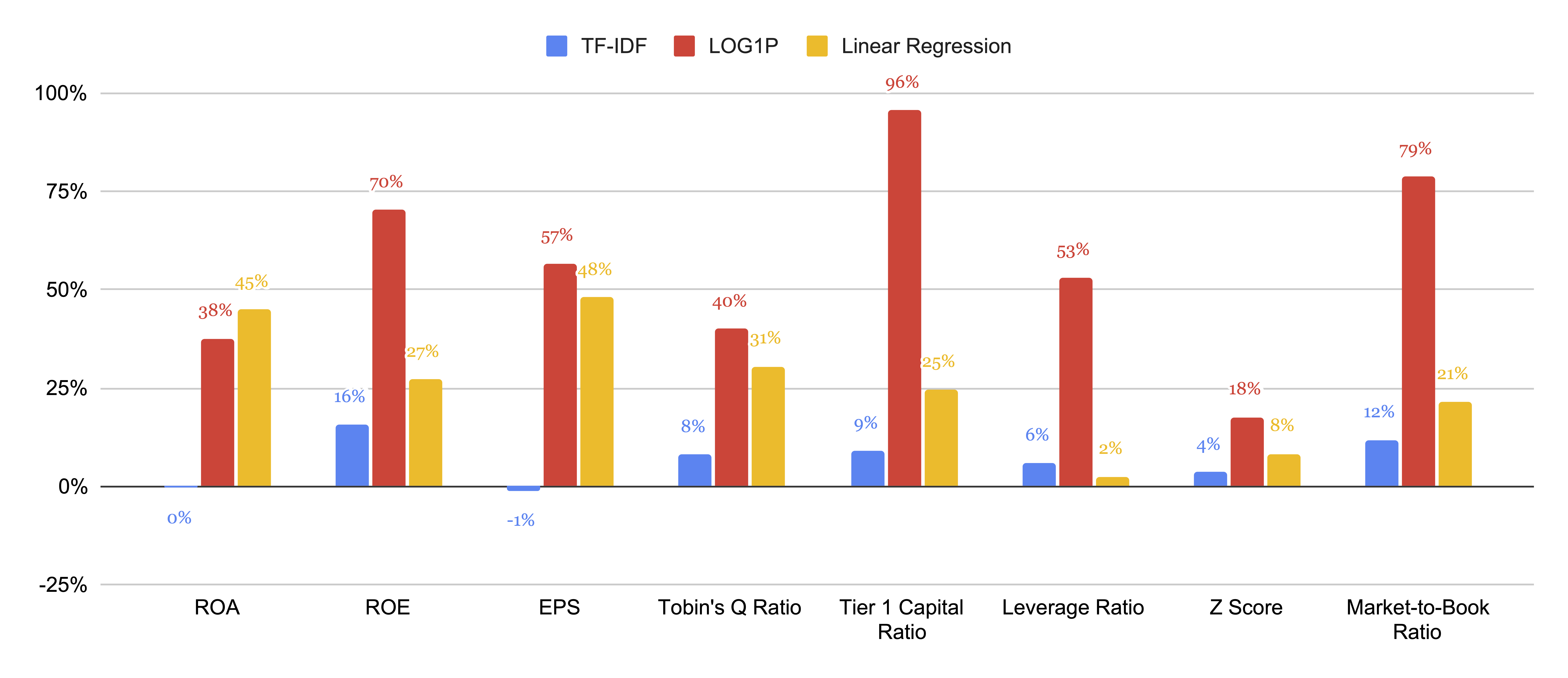}
    \vspace{-0.2in}
\caption{FETILDA Improvement on Item 1A over TF-IDF, LOG1P, and linear regression.}
\label{fig:sec1Ahistchart}
\end{figure}

To see the percentage improvements, in Figure \ref{fig:sec1Ahistchart}, for
every metric, we select the best performing version of FETILDA with frozen
layers in terms of mean squared error and calculate its improvement against
the text and numeric baseline methods. We can observe that our approach
outperforms LOG1P+ and linear regression on all of the metrics, and
outperforms TF-IDF on six of the metrics, the exceptions being ROA and EPS,
where TF-IDF has an extremely slight advantage.

\begin{table}[!ht]\centering
\caption{MSE Results on FIN10K. Best results in bold.}
\label{tab:tsai}
\resizebox{\linewidth}{!}{%
\begin{tabular}{|l|rrrrrr|r|}
\hline
Model\textbackslash Year &2001 &2002 &2003 &2004 &2005 &2006 & \textbf{Average} \\
\hline
BL (SVR) &0.174700 &0.160020 &0.187340 &0.144210 &0.136470 &0.146380 &0.150860 \\
LOG1P+: ALL &0.180820 &0.171750 &0.171570 &0.128790 &0.130380 &0.142870 &0.154360 \\
LOG1P+: SEN &0.185060 &0.163670 &\textbf{0.157950} &0.128220 &0.130290 &0.139980 &0.150860 \\
TF-IDF: ALL &0.123816 &0.121450 &0.218520 &0.176087 &0.148645 &0.138113 &0.154438 \\
\hline
FETILDA w/BERT (Fully Unfrozen) &0.128406 &0.111145 &0.180670 &0.111339 &0.094401 &0.091456 &0.119569 \\
FETILDA w/FinBERT (Fully Unfrozen) &\textbf{0.090408} &\textbf{0.108134} &0.172562 &\textbf{0.106124} &\textbf{0.090766} &\textbf{0.088401} &\textbf{0.109399} \\
FETILDA w/Longformer (Fully Unfrozen) &0.124797 &0.109595 &0.183509 &0.113019 &0.094623 &0.090408 &0.119325 \\
\hline
FETILDA w/BERT (Last Layer Frozen) &0.129132 &0.111559 &0.181691 &0.110962 &0.093300 &0.089595 &0.119373 \\
FETILDA w/FinBERT (Last Layer Frozen) &0.125969 &0.109420 &0.176483 &0.108349 &0.092103 &0.089228 &0.116925 \\
FETILDA w/Longformer (Last Layer Frozen) &0.135215 &0.114627 &0.193750 &0.117404 &0.096162 &0.089970 &0.124521 \\
\hline
FETILDA w/BERT (Fully Frozen) &0.121354 &0.108529 &0.175446 &0.108837 &0.093004 &0.090500 &0.116278 \\
FETILDA w/FinBERT (Fully Frozen) &0.118620 &0.113750 &0.159487 &0.108527 &0.097878 &0.095545 &0.115635 \\
FETILDA w/Longformer (Fully Frozen) &0.126380 &0.109627 &0.169686 &0.108116 &0.091884 &0.089902 &0.115932 \\
\hline
FETILDA w/Nystromformer &0.120945 &0.108224 &0.174019 &0.109716 &0.095050 &0.093098 &0.116842 \\
\hline

\end{tabular}}
\end{table}



\paragraph{FIN10K Dataset}\label{paragraph:tsaicomp} 
Table \ref{tab:tsai} compares the
performance of our method with the baseline method and the two versions of
the LOG1P+ model used in Tsai and Wang \cite{tsai2017risk}, and TF-IDF. 
The baseline method BL is essentially support vector
regression using the logarithm of the historic volatility for the prior
twelve months, also from \cite{tsai2017risk}. LOG1P+: ALL refers to the
model trained on the entire original text using the LOG1P features. Finally,
LOG1P+: SEN refers to the model trained on only the sentiment bearing words
taken from the L\&M dictionary~\cite{loughran2011liability}. We report the
results for these methods directly from their paper~\cite{tsai2017risk}. We
also include the results for the TF-IDF baseline. Among their methods,
LOG1P+: SEN performs the best for all years, except 2001.
For the average,
LOG1P+:SEN also performs better than TF-IDF, even though TF-IDF performs
better than LOG1P+:SEN for three out of the six years. However, as we can
observe, with the exception of 2003, FETILDA outperforms LOG1P+:SEN by a
large margin. Interestingly, on this much larger dataset, FETILDA w/FinBERT
outperforms both BERT and Longformer on all the metrics. It is the best
performing model over all the years, with the exception of 2003. 
{\em Looking at
the last column, which shows the average performance across the years
2001-2006, FETILDA w/FinBERT is the best; it outperforms Longformer by 8.3\%
(which is the second best method on average) and outperforms LOG1P+:SEN by
27\%. It outperforms all previous baselines by a significant margin, establishing new SOTA results.} 

\begin{figure}[!ht]
    \centering
    \includegraphics[scale = 0.035]{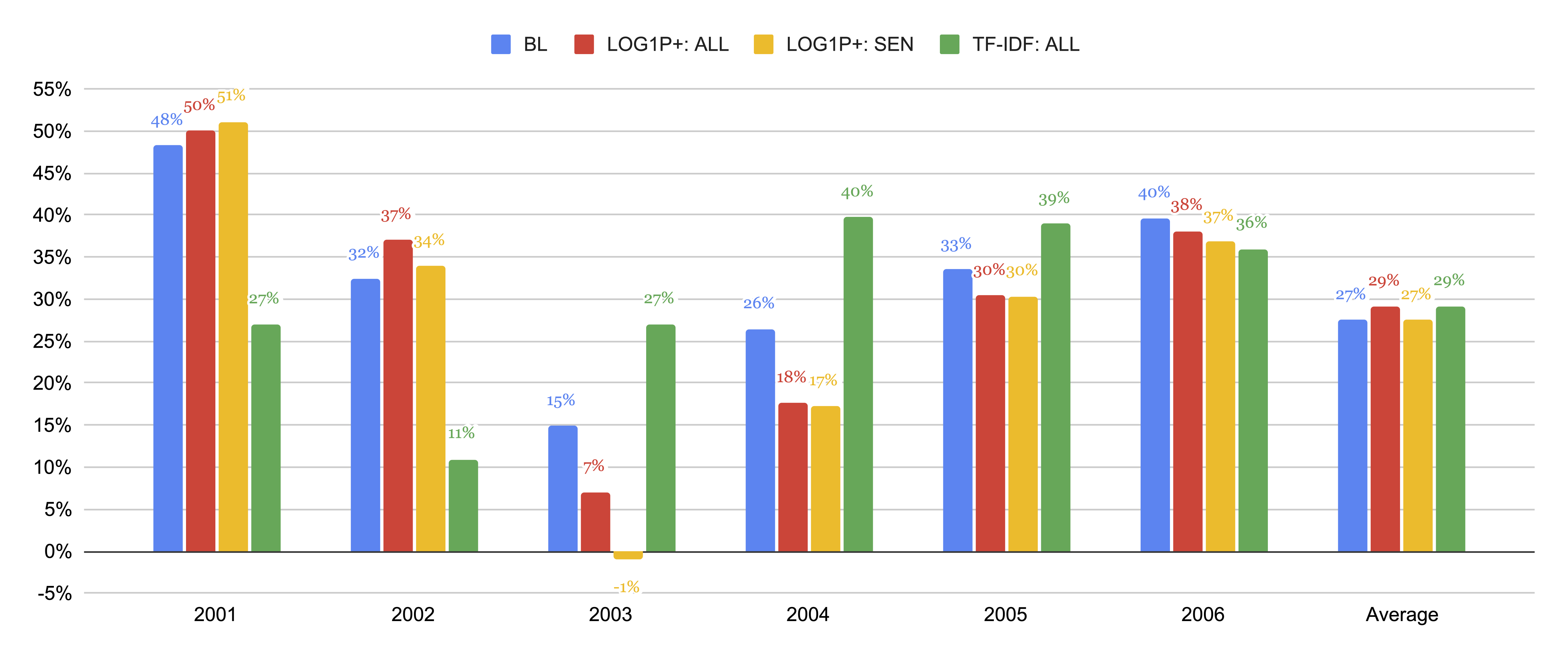}
    \caption{Best improvements of our approach compared with the LOG1P+ methods used in Tsai and Wang \cite{tsai2017risk}.}
    \label{fig:tsaicomp}
\end{figure}

Figure \ref{fig:tsaicomp} quantifies the percentage improvement of the best
performing version of FETILDA, in terms of mean squared error for volatility
prediction, versus the three approaches in Tsai and Wang
\cite{tsai2017risk}, as well as against TF-IDF baseline. Our approach
outperforms, BL, TF-IDF and LOG1P+:ALL/SEN by a significant margin, on
average (last set of bars) over 27\% across the years. These results clearly
showcase the benefits of contextual modeling of long text documents compared
to using simpler textual features based on term counts as done in LOG1P+.

Table~\ref{tab:tsai} also shows what happens to the FETILDA variants if we
freeze the layers of the language model and used only the pre-trained
embeddings, compared to fine-tuning through unfreezing all the layers or
only freezing the last layer. Interestingly, for the larger FIN10K dataset,
fine-tuning results in a much better model for FinBERT, only losing to the
fully frozen FinBERT in 2003, but not so much for BERT and Longformer.
Furthermore, the domain-specific pre-training in FinBERT followed by
fine-tuning results in the best overall model. On average, fine-tuned
FinBERT outperforms the frozen FinBERT by 5.4\%.

\subsection{Algorithmic Choices} 
Having show the effectiveness of our
FETILDA framework, we now present some results to justify some of the
algorithmic choices, such as which chunk-level pooling strategy does the
best and which FinBERT model performs the best. 

\begin{table}[!ht]
\centering
\caption{A comparison of three different models of FinBERT.}
\label{tab:3finberts}
\begin{tabular}{|l|r|r|r|r|}
\hline
Results\textbackslash Models &Araci\cite{Araci} &DeSola et al.\cite{desolafinbert} &Yang et al.\cite{Yang} \\
\hline
Validation loss &0.0011482 &0.0010539 &\textbf{0.0010205} \\
Testing error &0.0007781 &0.0008682 &\textbf{0.0007458} \\
\hline

\end{tabular}
\end{table}

\paragraph{FinBERT Variants} 
As discussed in related work, there are four
different FinBERT approaches proposed recently. Out of these, the
implementation for Liu et. al FinBERT \cite{Liu} is not publicly available.
We therefore compare the three FinBERT implementations that are available:
Araci \cite{Araci}, DeSola \cite{desolafinbert}, and Yang et al.
\cite{Yang}. Table \ref{tab:3finberts} shows the MSE results when predicting
ROA using both textual data from Item 7/7A and numeric historic data (using
a learning rate of 0.001) for the US Banks dataset. The results show that
Yang et al. implementation results in the best performance. We thus choose
the Yang et. al FinBERT \cite{Yang} as the underlying FinBERT model for
FETILDA. Recall that this FinBERT model was pre-trained on a very huge
financial corpus containing 10-K and 10-Q reports, earnings call
transcripts, and analyst reports.

\begin{table}[!ht]
\centering
\caption{A comparison of different pooling methods.}
\label{tab:diffpool}
\small
\resizebox{\linewidth}{!}{%
\begin{tabular}{|l|rr|rr|r|r|}
\hline
Results\textbackslash Methods &\multicolumn{2}{c|}{Mean pooling} &\multicolumn{2}{c|}{Max pooling} & Default pooling  \\
&Second-to-last layer &Last four layers &Second-to-last layer &Last four layers & Last layer\\
\hline
Validation MSE &0.0011465 &0.0012064 &0.0011102 &0.0011188 &\textbf{0.0010205} \\
Testing MSE &0.0008547 &0.0008221 &0.0007686 &0.0008820 &\textbf{0.0007458} \\
\hline

\end{tabular}}
\end{table}

\paragraph{Chunk-level Pooling Strategy} 
Recall that in
subsection~\ref{sec:3.2} we outlined several chunk-level pooling strategies
to create the final chunk embeddings. These include: (1) the default pooling
method (default pooler output) using the hidden state of the first token of
the last layer, (2) mean pooling method using the hidden states of the
second-to-last layer, (3) mean pooling method using the hidden states of the
last four layers, (4) max pooling method using the hidden states of the
second-to-last layer, and (5) max pooling method using the hidden states of
the last four layers. In Table \ref{tab:diffpool}, we present the
comparative MSE results for these alternatives on Item 7/7A for predicting
ROA. We observe that the default pooler output yields the best results for
both validation and testing datasets. We thus chose the default pooling
method using the hidden state of first token of the last layer, and this is
used for the different versions of FETILDA in our experiments above.

\section{Conclusion}\label{sec5}
In this paper, we presented our novel FETILDA framework to address the main
challenge of creating effective document embeddings for very long financial
text documents, such as 10-K public disclosures to the SEC, for which just
one section, such as Item 7/7A, contains over 12000 words on average. Even
the SOTA language models struggle to create informative document
representations for downstream tasks. Our FETILDA framework divides the long
documents into smaller chunks, and first learns chunk-level contextual
embeddings using SOTA language models like BERT, Longformer, and
Nystromformer, as well as domain-specific LMs like FinBERT. Next, we propose
a Bi-LSTM layer with self-attention to pool together the chunk embeddings
into the final document level embedding that weights different chunks based
on the attention scores. We apply FETILDA to the task of predicting eight
different KPIs for US Bank performance, as well as stock volatility
prediction for US companies from FIN10K. Our approach is shown to outperform
previous baselines, yielding SOTA results on the various regression tasks
for the two datasets used. With the FIN10K dataset especially, we
demonstrated quite evidently the significance of the improvement we get from
taking a domain-specific LM such as FinBERT and fine-tuning it on our
particular downstream task. We show this not only by how much
FETILDA with fully unfrozen FinBERT outperforms the baseline methods, but
also by how fine-tuning FinBERT through unfreezing all its layers during
training yields better performance than using the frozen pretrained
embeddings that the LM produces.

Our work opens avenues for follow-on research. For example, while the
contextual models in FETILDA can learn more effective document
representations compared to baselines like TF-IDF, there is still scope for
more improvement. For instance, we found that the performance is better on
longer documents, but the language models lose some edge on shorter
documents (e.g., those from Item 1A). We plan to explore this in more detail
on even larger 10-K datasets to confirm the trends. 
One could also consider learning
even larger domain-specific pre-trained models for financial text, with
larger blocks (e.g., using Longformer or Nystromformer instead of BERT for
pre-training). Finally, we plan to explore alternative approaches to learn
better document representations. For example, instead of using the entire
text, we can focus on the most important words, phrases, and sentences
(e.g., sentiment bearing elements within the text). We can derive better
chunk-level and document-level embeddings in this manner. How to select
these informative elements from text remains an open challenge.










\bibliographystyle{unsrt}  
\bibliography{references}

\end{document}